%% file: main.tex
\documentclass[11pt]{article}

\usepackage[letterpaper,margin=0.8in]{geometry}
\usepackage[utf8]{inputenc}
\usepackage{lmodern}
\usepackage{authblk}
\usepackage[numbers,sort&compress]{natbib}

\usepackage[T1]{fontenc}
\usepackage{graphicx}
\usepackage{amsmath}
\usepackage{amssymb}
\usepackage{booktabs}
\usepackage{multirow}
\usepackage{pifont}
\usepackage[table]{xcolor}
\usepackage{subcaption}
\usepackage{tabularx}
\usepackage{siunitx}
\usepackage{url}
\usepackage[hidelinks]{hyperref}
\usepackage{orcidlink}

\usepackage{etoolbox}
\usepackage{adjustbox}
\AtBeginEnvironment{table*}{%
  \let\origtabular\tabular \let\endorigtabular\endtabular
  \renewenvironment{tabular}%
    {\adjustbox{max width=\linewidth}\bgroup\origtabular}%
    {\endorigtabular\egroup}}

\newcommand{\wdtwo}{WildDepth-v2}

\newif\ifdraft \drafttrue
\ifdraft

  \newcommand{\todo}[1]{\textcolor{magenta}{\textbf{[TODO: #1]}}}
\else

  \newcommand{\todo}[1]{}
\fi

\newcommand{\figph}[3]{%
  \IfFileExists{#1}{\includegraphics[width=#2]{#1}}{%
    \fbox{\parbox[c][0.45#2][c]{\dimexpr#2-2\fboxsep-2\fboxrule\relax}{\centering\footnotesize
      \textcolor{red}{\textbf{FIGURE PLACEHOLDER}}\\[2pt]\texttt{\detokenize{#1}}\\[2pt]#3}}}}

\usepackage{xcolor}
\usepackage{fontawesome5}   
\definecolor{wilddepthblue}{HTML}{1F77B4} 

\title{WildDepth-v2: A Synchronized RGB--LiDAR--Radar Dataset and Benchmark for Metric Depth, Motion, and Tracking of Wildlife}

\author[1*]{Muhammad Aamir\orcidlink{0000-0002-4999-7740}}
\author[1]{Sangyun Shin\orcidlink{0000-0003-0766-4101}}
\author[2]{Naoya Muramatsu\orcidlink{0000-0002-8976-2292}}
\author[3]{Matthew Wijers\orcidlink{0000-0002-5097-8758}}
\author[4]{Amir Patel\orcidlink{0000-0002-2344-4179}}
\author[3]{Andrew Loveridge\orcidlink{0000-0002-3493-6419} }
\author[1]{Andrew Markham\orcidlink{0000-0001-5716-3941}}

\affil[1]{Department of Computer Science, University of Oxford, Oxford, UK}
\affil[2]{University of Cape Town, Cape Town, South Africa}
\affil[3]{Wildlife Conservation Research Unit, Department of Biology, University of Oxford, UK}
\affil[4]{Department of Computer Science, University College London, UK}

\date{}

\begin{document}

\maketitle

\begin{center}
\href{https://yunshin.github.io/WildDepth/}{\textcolor{wilddepthblue}{\underline{\textbf{\faGlobe\ Project Page}}}}
\end{center}

\input{sections/0_abstract}
\input{sections/1_introduction}

\input{sections/2_related_work}
\input{sections/3_dataset}

\input{sections/4_benchmark}
\input{sections/5_experiments}
\input{sections/6_discussion}
\input{sections/7_conclusion}

\bibliographystyle{plainnat}
\bibliography{egbib,v2_additions}

\end{document}

%% file: sections/0_abstract.tex
\begin{abstract}
Measuring where an animal is, how far it is, and how it moves, in metres and metres per second rather than pixels -- underpins distance-sampling abundance estimates, morphometrics, and movement ecology.
Monocular video, the dominant wildlife data source, carries no metric scale, and existing animal datasets provide no metric ground truth against which to test methods that try to recover it.
We previously introduced \emph{WildDepth}, a synchronized RGB--LiDAR wildlife dataset from three sites on two continents.
Here we present \textbf{\wdtwo}, which extends it in three directions.
(i)~\emph{A third sensing modality}: 37 sequences (19{,}716 synchronized frames) of RGB--LiDAR--4D-imaging-radar recordings of animals, with per-point Doppler velocity and a measured error budget, to our knowledge the first wildlife data captured with imaging radar.
(ii)~\emph{A frozen, held-out benchmark}: a checksummed hidden test set for RGB-only distance estimation and RGB+sparse-LiDAR depth completion, scored separately on animal and background pixels.
(iii)~\emph{Motion and tracking tasks}: metric 3D animal trajectories obtained by lifting 2D tracks with LiDAR, radar, or monocular depth. On 48{,}723 animal frames, object-level radar range calibrates a monocular depth model better than scene-level LiDAR calibration and without LiDAR at inference (AbsRel 0.120 vs.\ 0.198): note however LiDAR at inference on the animal remains more accurate (0.052).
Over 431 animal tracks, scaling monocular depth by radar once per track reduces 3D position error four-fold (2.20$\rightarrow$0.58\,m) and the bias of a distance-sampling detection function from $-$13\% to $-$2\%, matching a LiDAR calibration made once per track.
Doppler velocity, transferred onto the co-registered LiDAR, yields motion masks 3.7$\times$ denser than radar alone at 72\% precision, is able to gate multi-second LiDAR accumulation on moving animals ($-$37\% error), and tracks radial animal speed (r\,=\,0.60), although it under-reads it by half.
We also map radar's limits: a 1.8\,m vertical versus 0.1\,m in-plane extrinsic residual rules out direct per-pixel radar depth at this point.
Data, calibration, benchmark, and code are released.
\end{abstract}

%% file: sections/1_introduction.tex
\section{Introduction}
\label{sec:intro}

Non-invasive monitoring of animals is increasingly vital as biodiversity faces unprecedented pressure from habitat loss, poaching, and climate change~\cite{ceballosBiologicalAnnihilationOngoing2017,tuiaPerspectivesMachineLearning2022}.
Computer vision has made detection, species classification, and individual identification routine at scale~\cite{beeryMegadetector2019,gadotCropNotCrop2024,valIdentifyingOcelotsIndividual2024}, and multi-object tracking now links detections into trajectories in video~\cite{zhang2022bytetrack,romero2019idtracker,mathis2018deeplabcut}.
Many of the questions ecologists ask of these trajectories, however, are \emph{metric}: how far away is the animal (the input to distance-sampling abundance estimates~\cite{buckland2001distance,howe2017distance}), how large is it (body condition, sexual dimorphism, population structure~\cite{daviesAdvancesAnimalEcology2014}), and how fast and where does it move (behaviour, energetics).
The overwhelming majority of wildlife imagery -- camera traps, hand-held and aerial video~\cite{beeryMegadetector2019,gabeff2025mammalps,lila_science_2026} -- is monocular RGB and carries no metric scale.
Modern monocular depth estimators (MDEs) recover plausible relative geometry, but whether their metric outputs can be trusted for a distant, deformable animal in unstructured terrain has been difficult to test, because animal datasets do not contain metric ground truth.

In the conference version of this work~\cite{aamirWildDepth2026} we introduced \emph{WildDepth}, a synchronized RGB--LiDAR dataset collected at three ecologically distinct sites including Kgalagadi Transfrontier Park (South Africa), Bubye Valley Conservancy (Zimbabwe), and Longleat Safari Park (UK), and used it to show that off-the-shelf MDEs degrade sharply at range while sparse LiDAR anchoring roughly halves their metric error.
LiDAR, however, is expensive, power-consuming, and mechanically delicate, which limits how widely such a rig can be deployed; and neither LiDAR nor a single RGB frame measures \emph{motion} directly, which is exactly the quantity a tracker must reason about.

Millimetre-wave 4D imaging radar offers a complementary trade-off that has transformed automotive perception~\cite{caesar2020nuscenes,palffy2022multi,paek2022kradar}: it is inexpensive, low-power, robust to dust, rain and darkness, and it measures each return's \emph{radial velocity} through the Doppler effect.
Radar--camera fusion for depth, detection, and tracking is now an active area in driving~\cite{long2021radar,lin2020depth,nabati2021centerfusion,yang2020radarnet}, but the targets there are large, rigid, and retro-reflective.
Wildlife inverts these conditions: animals possess small, low-reflectivity, non-rigid bodies whose Doppler signature is dominated by limb motion, observed at tens of metres superimposed on natural clutter.
Whether radar transfers to this regime is, to our knowledge, untested.

This paper, \textbf{WildDepth-v2}, extends WildDepth to address both gaps, a cheaper source of metric scale, and a direct measurement of motion and transforms the dataset into a benchmark for metric animal perception over time.
Our contributions are:
\begin{itemize}
  \item[] \textbf{The first RGB--LiDAR--radar wildlife recordings} (Sec.~\ref{sec:radar_data}). We add a synchronized Altos~V2 4D imaging radar to the Longleat subset: 37 sequences and 19{,}716 frames in which RGB, LiDAR, and radar are all present, with per-point Doppler velocity and radar cross-section. We release the radar$\rightarrow$LiDAR extrinsic together with a \emph{measured} error budget (radar--LiDAR timing, and a strongly anisotropic extrinsic residual), rather than assuming one.
  \item[] \textbf{A frozen, held-out WildDepth benchmark} (Sec.~\ref{sec:benchmark}). A checksummed hidden test set drawn from unreleased footage, with two tracks, first RGB-only camera-to-animal distance and RGB+sparse-LiDAR dense depth and second a scorer that separates animal from background error, and a documented construction recipe and defect list.
  \item[] \textbf{Radar as a metric anchor and a motion sensor} (Secs.~\ref{sec:exp_radar_scale}--\ref{sec:exp_motion_dens}). Object-level radar range calibrates monocular depth better than scene-level LiDAR calibration without LiDAR at inference (AbsRel 0.120 vs.\ 0.198 on 20{,}869 test animal frames), though not as well as LiDAR on the animal. Doppler, transferred to the co-registered LiDAR through anisotropic association, yields motion masks 3.7$\times$ denser than radar alone at 72\% precision, and can gate multi-second LiDAR accumulation so that moving animals are not smeared.
  \item[] \textbf{Metric motion and tracking tasks} (Sec.~\ref{sec:exp_tracking}). We lift 2D animal tracks to metric 3D trajectories using LiDAR, radar, or monocular depth and evaluate position, speed, jitter and distance-sampling bias. Scaling monocular depth by radar once per track matches a once-per-track LiDAR calibration (0.58 vs.\ 0.52\,m position error) and cuts the detection-function bias of monocular depth from $-$13\% to $-$2\%.
  \item[] \textbf{An objective map of radar's operating envelope for wildlife} (Sec.~\ref{sec:exp_envelope}), by including negative results we believe are useful to others: per-pixel radar depth fails because of a 1.8\,m vertical extrinsic residual; radar does not extend coverage or see through vegetation in our data; Doppler tracks radial animal speed but under-reads it by half; and per-frame extrinsic refinement can drift far enough to corrupt LiDAR ground truth.
\end{itemize}

Relative to the conference paper~\cite{aamirWildDepth2026}, the radar modality, the hidden benchmark and its baselines, the motion and tracking tasks, and the envelope analysis are new; the monocular-depth evaluation has been recomputed on the publicly released snippets with seven models (Sec.~\ref{sec:exp_mde}); and the behaviour and 3D-reconstruction baselines are summarized for completeness (Sec.~\ref{sec:exp_eccv_tasks}).

%% file: sections/2_related_work.tex
\section{Related Work}
\label{sec:related}

\noindent\textbf{Wildlife datasets.}
Camera traps, aerial surveys, and hand-held video have produced large animal datasets for detection, identification, and behaviour recognition~\cite{ravoor2020deep,beeryMegadetector2019}.
MammAlps~\cite{gabeff2025mammalps}, LoTE~\cite{liu2023lote}, KABR~\cite{kholiavchenko2024kabr} and BaboonLand~\cite{duporge2025baboonland} provide video benchmarks for behaviour; Animal Kingdom~\cite{ng2022animal} and MammalNet~\cite{chen2023mammalnet} capture species appearance from internet video; ChimpACT~\cite{ma2023chimpact} and the Meerkat dataset~\cite{rogers2023meerkat} target tracking and behaviour of a single species.
These datasets are RGB-only (occasionally with audio) and contain no metric geometry.
WildPose~\cite{muramatsuWildPoseLongRange3D2024} demonstrated long-range RGB--LiDAR capture of wild animals, and WildDepth~\cite{aamirWildDepth2026} turned it into a multi-site dataset.
\wdtwo{} adds imaging radar and a held-out benchmark (Table~\ref{tab:literature}).

\noindent\textbf{Animal tracking.}
Tracking-by-detection with Kalman-filter motion models and IoU or appearance association, SORT~\cite{bewley2016sort}, DeepSORT~\cite{wojke2017deepsort}, ByteTrack~\cite{zhang2022bytetrack}, is the default for animals in video, alongside laboratory trackers for collectives~\cite{romero2019idtracker} and pose trackers~\cite{mathis2018deeplabcut,ye2024superanimal,mathisDeepLearningTools2020}.
In driving, tracking has moved to metric 3D space~\cite{weng2020ab3dmot,yin2021centerpoint} and is scored with CLEAR-MOT and HOTA~\cite{bernardin2008clearmot,luiten2021hota}.
Animal tracking in the wild remains almost entirely 2D because metric 3D ground truth has been unavailable; \wdtwo{} provides it through LiDAR and adds per-point radial velocity from radar.

\noindent\textbf{Monocular depth estimation.}
MDE has advanced from self-supervised CNNs~\cite{li2024improved} to large transformers DPT~\cite{ranftl2021vision}, Depth Anything~\cite{yang2024depth}, Depth Pro~\cite{bochkovskii2024depth}, and metric geometry models such as MoGe-2~\cite{wangMoGe2AccurateMonocular2025} evaluated mainly on driving and indoor benchmarks~\cite{liao2022kitti,silberman2012indoor}.
Wildlife scenes are unstructured, their subjects are non-rigid and often tens of metres away, and they are under-represented in training data; recent work has begun lifting monocular animal observations to metric 3D~\cite{shuklaWildLIFTLiftingMonocular2026,aamirRobustMetricDistance2025,Zuffi_2024_CVPR}.
We provide the metric ground truth to test such methods and study which sparse sensor can supply the missing scale.

\noindent\textbf{Radar for depth, motion, and tracking.}
Automotive radar datasets such as nuScenes~\cite{caesar2020nuscenes}, View-of-Delft~\cite{palffy2022multi}, K-Radar~\cite{paek2022kradar} enabled radar--camera fusion for depth completion~\cite{lin2020depth,long2021radar}, detection~\cite{nabati2021centerfusion}, and Doppler-aware tracking of dynamic objects~\cite{yang2020radarnet}.
Doppler also gives platform ego-motion~\cite{kellner2013instantaneous}, and micro-Doppler from limbs and rotating parts is a well-known signature of articulated targets~\cite{chen2006microdoppler}.
All of this work targets vehicles and pedestrians on roads.
Wildlife inverts the assumptions: small, low-reflectivity, deformable targets at long range, in natural clutter, with Doppler dominated by gait.
We are not aware of prior work that evaluates imaging radar on animals.

\begin{table*}[t]
\centering
\caption{Publicly available wildlife datasets. \wdtwo{} is the only one with metric depth, and the only one with imaging radar. Frame counts are curated frames.}
\label{tab:literature}
\begin{tabular}{lcccccc}
\toprule
\textbf{Dataset} & \textbf{Source} & \textbf{\# Species} & \textbf{\# Frames} & \textbf{RGB} & \textbf{LiDAR} & \textbf{Radar} \\
\midrule
AnimalKingdom~\cite{ng2022animal} & YouTube & 850 & 30k & \ding{51} & \ding{55} & \ding{55} \\
MammalNet~\cite{chen2023mammalnet} & YouTube & 173 & 20k & \ding{51} & \ding{55} & \ding{55} \\
KABR~\cite{kholiavchenko2024kabr} & Drone & 3 & 13k & \ding{51} & \ding{55} & \ding{55} \\
Meerkats~\cite{rogers2023meerkat} & Zoo & 1 & 848k & \ding{51} & \ding{55} & \ding{55} \\
ChimpACT~\cite{ma2023chimpact} & Zoo & 1 & 160k & \ding{51} & \ding{55} & \ding{55} \\
MammAlps~\cite{gabeff2025mammalps} & Camera traps & 5 & 6k & \ding{51} & \ding{55} & \ding{55} \\
LoTE~\cite{liu2023lote} & Camera traps & 11 & 10k & \ding{51} & \ding{55} & \ding{55} \\
BaboonLand~\cite{duporge2025baboonland} & Drone & 1 & 30k & \ding{51} & \ding{55} & \ding{55} \\
WildDepth~\cite{aamirWildDepth2026} & Camera/LiDAR & 29 & 202k & \ding{51} & \ding{51} & \ding{55} \\
\textbf{\wdtwo{} (ours)} & Camera/LiDAR/Radar & 29 & 202k$^\dagger$ & \ding{51} & \ding{51} & \ding{51}$^\ddagger$ \\
\bottomrule
\end{tabular}\\[3pt]
{\footnotesize $^\dagger$Plus a held-out hidden test set (Sec.~\ref{sec:benchmark}). $^\ddagger$19{,}716 RGB--LiDAR--radar frames from the Longleat subset (Sec.~\ref{sec:radar_data}).}
\end{table*}

%% file: sections/3_dataset.tex
\section{The \wdtwo{} Dataset}
\label{sec:dataset}

\begin{figure*}[t]
\centering
\figph{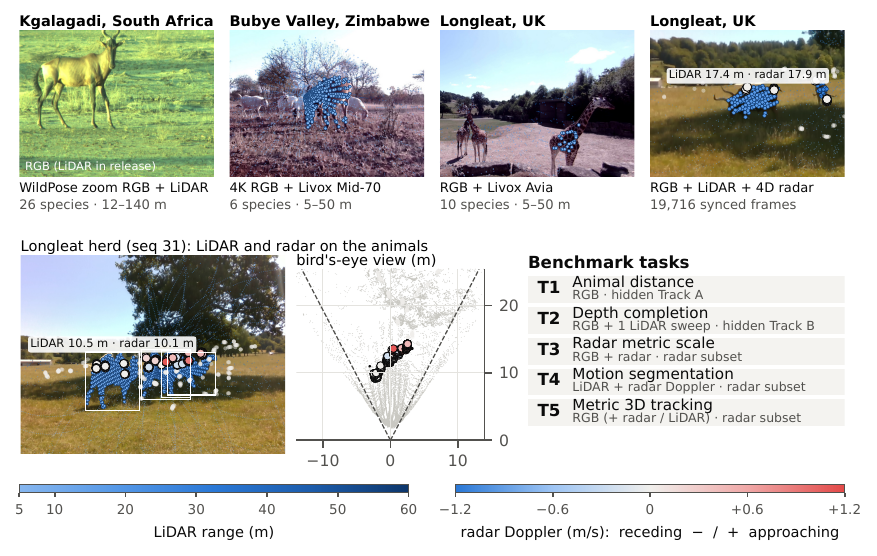}{\textwidth}{Overview: three sites (map), two Longleat rigs, sample RGB with projected LiDAR (colour = range) and radar (colour = Doppler) for one frame; arrows to the five tasks of Sec.~\ref{sec:benchmark}.}
\caption{\wdtwo{} overview. RGB--LiDAR recordings from three sites (left), a synchronized RGB--LiDAR--radar subset at Longleat in which each radar return carries range, azimuth, elevation, Doppler velocity and RCS (centre), and the benchmark tasks (right).}
\label{fig:overview}
\end{figure*}

\wdtwo{} consists of the RGB--LiDAR collection of WildDepth~\cite{aamirWildDepth2026} (Sec.~\ref{sec:rgb_lidar}), a new radar-equipped subset (Sec.~\ref{sec:radar_data}), and a new held-out test set (Sec.~\ref{sec:benchmark}).
Figure~\ref{fig:overview} gives an overview.

\subsection{RGB--LiDAR Collection (from WildDepth)}
\label{sec:rgb_lidar}

WildDepth comprises three subsets collected in 2022 and 2025 in wild habitats and a semi-controlled safari park (Table~\ref{tab:dataset_stats}).
The \textbf{Kgalagadi} subset (Kgalagadi Transfrontier Park, South Africa, December 2022) was captured over 13 days with the vehicle-mounted WildPose system~\cite{muramatsuWildPoseLongRange3D2024}, a zoom-lens camera with a Livox TELE-15 LiDAR, and covers 26 species, apex predators such as lion, cheetah and leopard, and herbivores such as springbok and gemsbok, at 12--140\,m.
The \textbf{Bubye} subset (Bubye Valley Conservancy, Zimbabwe, June 2025) used a custom 4K RGB camera with a Livox Mid-70, in mobile (vehicle window) and stationary configurations, over mopane and acacia woodland; it covers six species at 5--50\,m.
The \textbf{Longleat} subset (Longleat Safari Park, UK, July 2025) covers ten species in open enclosures at 5--50\,m, recorded with an RGB camera and a Livox Mid-70 LiDAR (first rig) and, in parallel, with a second vehicle-mounted rig that additionally carries the imaging radar described below.
The curated release totals 202{,}941 frames in species-balanced ``snippets'', and the full raw recordings are also released.

The LiDARs are prism-scanning time-of-flight sensors rated at $\approx$2\,cm ($1\sigma$) range precision that stays roughly constant over our operating distances~\cite{LivoxTELE15User,LivoxMid70User2021,LivoxAviaUser}; the WildPose system was independently validated at a median absolute error of 45.2\,mm at 160\,m~\cite{muramatsuWildPoseLongRange3D2024}.
LiDAR runs at 10\,Hz and the cameras at 30\,Hz; each LiDAR scan is paired with the camera frame nearest in time ($\pm$10\,ms).
Animals were detected with MegaDetector~\cite{beeryMegadetector2019}, tracked with ByteTrack~\cite{zhang2022bytetrack}, and every track was verified for species and behaviour.

\paragraph{Release format.}
Each snippet ships RGB frames, a projected LiDAR depth map per frame (16-bit PNG, 0 = no return), a video, and metadata with global frame indices; calibration (rectified intrinsics, distortion, LiDAR$\rightarrow$camera extrinsic) is released per record.
\emph{Depth units differ between records} -- millimetres for Longleat and centimetres for Bubye -- and are stated in each record's README; the benchmark bundle re-encodes all depth to millimetres.

\begin{table}[t]
\centering
\caption{\wdtwo{} statistics. A snippet is a sequence of consecutive frames of the same species (or group) in a consistent setting.}
\label{tab:dataset_stats}
\resizebox{\columnwidth}{!}{%
\begin{tabular}{lccccc}
\toprule
Subset & Species & Snippets & Frames & Range (m) & Radar \\
\midrule
Kgalagadi & 26 & 131 & 75{,}741 & 12--140 & \ding{55} \\
Bubye & 6 & 37 & 84{,}700 & 5--50 & \ding{55} \\
Longleat & 10 & 70 & 42{,}500 & 5--50 & partial \\
Longleat subset (new) & 10 & 37 & 19{,}716$^\ast$ & 5--55$^\dagger$ & \ding{51} \\
Hidden test (Sec.~\ref{sec:benchmark}) & -- & 13 scenes & 902 & 5--50 & \ding{55} \\
\bottomrule
\end{tabular}}\\[2pt]
{\footnotesize $^\ast$Frames in which RGB, LiDAR and radar are all present (22{,}348 radar frames in total). $^\dagger$Range of evaluated animals; radar returns extend beyond 200\,m.}
\end{table}

\subsection{RGB--LiDAR--Radar Subset (new)}
\label{sec:radar_data}

\begin{figure}[t]
\centering
\figph{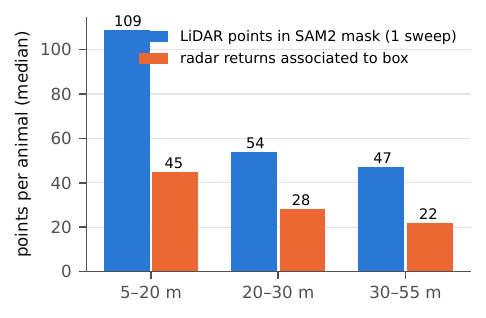}{0.7\columnwidth}{density}
\caption{Sensor density on animals, by ground-truth distance: single-sweep LiDAR points inside the SAM2 mask and radar returns associated with the detection box (median over 48{,}723 animals).}
\label{fig:density}
\end{figure}

\paragraph{Sensor.}
On the second rig we rigidly co-mounted an \textbf{Altos~V2 4D imaging radar} with the LiDAR and an Intel RealSense colour camera ($640{\times}480$).
Per frame, the radar returns a sparse point cloud in which each point carries its Cartesian position $(x,y,z)$, its \emph{radial (Doppler) velocity} $h$ in m/s, its signal strength (a radar-cross-section proxy, $s$), and a coarse three-way motion class $v\in\{-1,0,+1\}$; we recommend the continuous $h$, since the motion class under-counts moving returns by roughly an order of magnitude on our data.
Angular resolution is about $1.38^\circ$ in azimuth and $1.43^\circ$ in elevation.
A typical frame holds 679 valid returns (median over sequences) against 19{,}390 LiDAR points; returns extend beyond 200\,m, but the animals we study lie within 55\,m.
On animals, the radar returns associated with a detection (Sec.~\ref{sec:methods}) number a median of 45, 28 and 22 at 5--20, 20--30 and 30--55\,m, against 109, 54 and 47 single-sweep LiDAR points inside the animal's mask (Fig.~\ref{fig:density}); many of the associated radar returns come from ground and background seen past the animal. Radar is therefore much sparser on the animal than the LiDAR it accompanies, but it measures range and velocity directly.

\paragraph{Coverage.}
Radar was recorded on two days.
On the second day (July~8) it is healthy in all 37 snippets; on the first day (July~7) the sensor stopped publishing and the recorder repeated a stale message, so every radar frame within a snippet is byte-identical.
We detected this with a per-snippet checksum test and \emph{exclude} July~7 radar from the release and all experiments.
The radar subset thus contains 37 snippets, 22{,}348 radar frames, and 19{,}716 frames with RGB, LiDAR and radar all present (one LiDAR file is empty and excluded).

\paragraph{Synchronization.}
All three streams share one global 10\,Hz frame index, so radar, LiDAR and RGB with the same index form a synchronized triple; the index is also the key that links the second rig recordings to the released Longleat snippets.
Radar messages are matched to the nearest LiDAR timestamp, and we release the per-frame offset.
The mean absolute radar--LiDAR offset is 16.8\,ms (95th percentile 31.5\,ms), with rare dropouts up to 350\,ms that we flag (Fig.~\ref{fig:calib}); at a walking speed of 5\,m/s a 20\,ms offset corresponds to 10\,cm of displacement.

\paragraph{Calibration and its error budget.}
The LiDAR$\rightarrow$camera extrinsic of the second rig rig was solved once per sequence (the \emph{anchor}) and then refined per frame by joint correspondence and gradient optimization (intrinsics frozen). The per-frame refinement drifts: by a median of 2.1$^\circ$ from the anchor and by up to 25$^\circ$ in some sequences, which is implausible for a rigid mount. An independent check -- the spread of LiDAR depth inside SAM2 animal masks, which grows when background leaks into the mask -- favours the anchor wherever the drift exceeds 5$^\circ$ (in-mask IQR 0.83 vs.\ 1.27\,m; 116 vs.\ 50 LiDAR points on the animal), and the two agree below 1$^\circ$. We therefore use the anchor extrinsic for all ground truth. The depth maps shipped with the public second rig record were projected with the per-frame extrinsics and are affected in the high-drift sequences; we release re-projected maps with \wdtwo. 

The radar$\rightarrow$LiDAR extrinsic is a single rigid transform, because the two sensors share one mount; it was initialized by aligning static structure and refined manually on one sequence.
We measured its residual on 827k static radar returns across all 37 sequences, matching each to the nearest accumulated-LiDAR point in the ground plane, and found it strongly \textbf{anisotropic}: a median of 0.10\,m in range and azimuth (75th percentile 0.26\,m) but \textbf{1.84\,m in elevation} (Fig.~\ref{fig:calib}), a direct consequence of the radar's few elevation bins.
This anisotropy governs every radar experiment in this paper: projecting a radar return into the image can move it by several animal heights vertically, so we always associate radar to an animal at the \emph{object} level, the robust median range of the radar returns whose projection falls in the animal's box and never per pixel. We release the extrinsic, the residual analysis, and the association code.

\begin{figure*}[t]
\centering
\figph{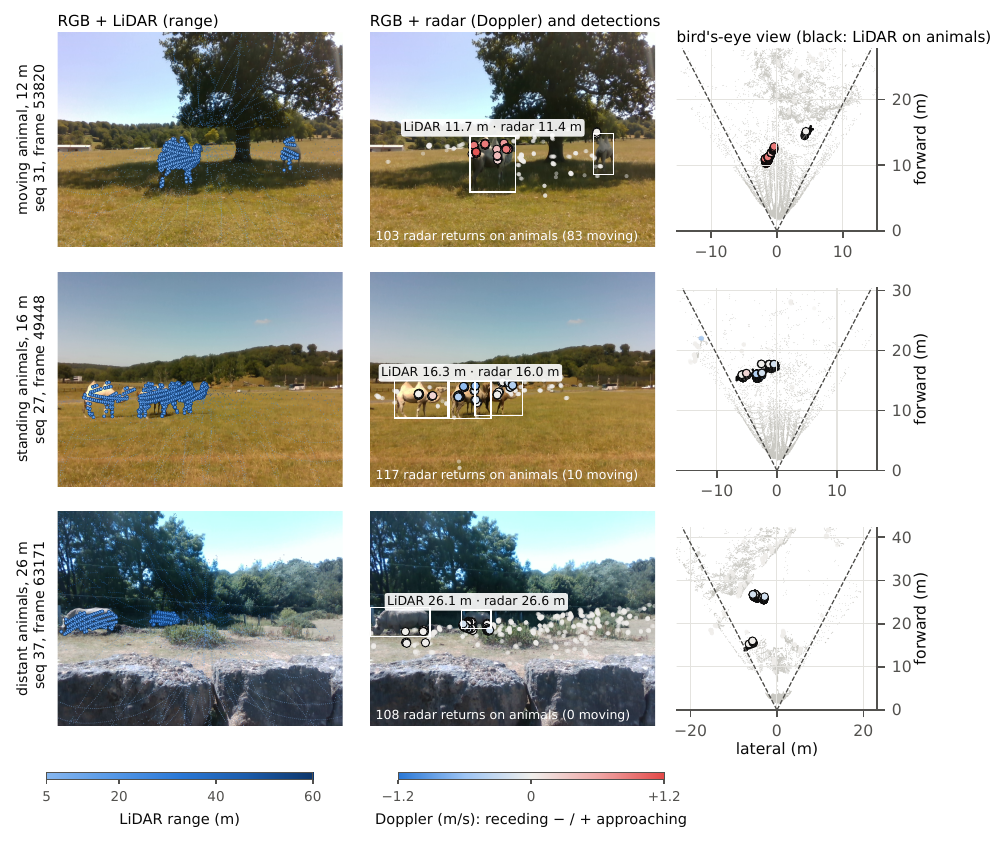}{\textwidth}{3--4 example frames (e.g.\ snippet\_28 f50195, snippet\_27 f49205): RGB; LiDAR projected (range colour); radar projected (Doppler colour, showing vertical offset from elevation error); BEV of radar+LiDAR.}
\caption{RGB--LiDAR--radar examples. Radar returns (coloured by Doppler velocity) sit on the animals in range and azimuth but are displaced vertically when projected into the image, reflecting the 1--2\,m elevation uncertainty of the radar extrinsic.}
\label{fig:radar_qual}
\end{figure*}

\paragraph{Ethics and privacy.}
All recordings were made from public roads or viewing platforms under permission of the site operators, without interacting with the animals.

%% file: sections/4_benchmark.tex
\section{Tasks, Benchmark, and Baseline Methods}
\label{sec:benchmark}

We define five tasks (Table~\ref{tab:tasks}).
T1 and T2 are scored on a frozen hidden test set; T3--T5 use the radar subset.

\begin{table}[t]
\centering
\caption{\wdtwo{} tasks. GT = ground truth.}
\label{tab:tasks}
 \resizebox{\columnwidth}{!}{%
\begin{tabular}{llll}
\toprule
Task & Input & Output & GT / data \\
\midrule
T1 Animal distance & RGB & metric depth map & LiDAR in mask; hidden A \\
T2 Depth completion & RGB + 1 LiDAR sweep & metric depth map & 11-sweep LiDAR; hidden B \\
T3 Radar metric scale & RGB + radar & metric distance & LiDAR in mask; radar subset \\
T4 Motion segmentation & LiDAR + radar & moving points & RGB boxes / flow; radar subset \\
T5 Metric 3D tracking & RGB (+LiDAR/radar) & 3D tracks, speed & LiDAR tracks; radar subset \\
\bottomrule
\end{tabular}}
\end{table}

\subsection{Hidden Test Set (T1, T2)}
\label{sec:hidden}

A benchmark built from released data invites overfitting, so we built a hidden test set from \emph{unreleased} Longleat MorphoCam footage recorded between released snippets.
Construction, in order: (1)~exclude every released snippet range with a guard band of $\pm$300 frames; (2)~keep only frames where the rig is static, measured by phase correlation against a running reference ($\leq$0.2\,px at 1/8 scale), so that LiDAR can be accumulated; (3)~accumulate 11 LiDAR sweeps as dense ground truth; (4)~detect animals with MegaDetector (confidence $\geq$0.5) and segment them with SAM2~\cite{raviSAM2Segment2024}; (5)~keep detections whose ground-truth distance lies in 5--50\,m; (6)~drop masks whose in-mask LiDAR depth has an inter-quartile range above 3\,m, which removes detections corrupted by background leakage. Two scenes with a fence across the lens (121 frames) were removed entirely. 

\begin{figure}[t]
\centering
\figph{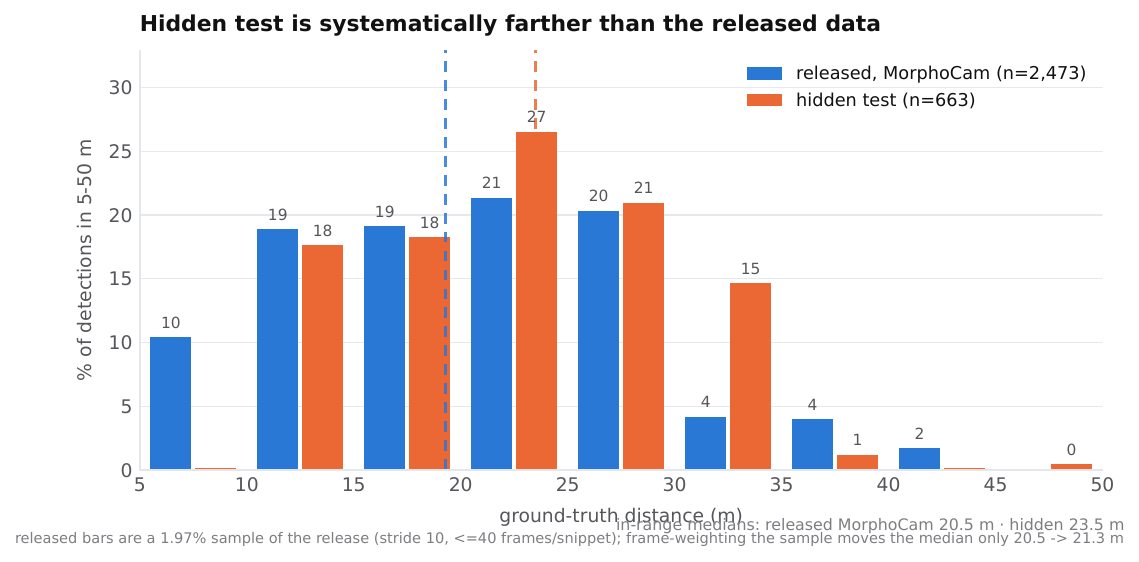}{\columnwidth}{Distance distribution of released vs hidden-test detections (already generated: depth\_eval/out/charts/chart\_released\_vs\_hidden.pdf)}
\caption{Ground-truth camera-to-animal distance of released-snippet detections versus the hidden test set.}
\label{fig:hidden_dist}
\end{figure}

The result has 13 scenes, 352 frames with 510 animal detections for \textbf{Track~A} (T1), and 902 frames for \textbf{Track~B} (T2), plus a public validation split (137 frames, 298 detections).
The set is frozen: an SHA-256 manifest covers all 4{,}046 files, the scorer refuses to run if the reference no longer matches it, and two independently verified copies are kept.
The current version (v1.3) merges the sparsely populated 5--10\,m bin into 10--20\,m (Fig.~\ref{fig:hidden_dist}).

\paragraph{Known limitations, stated up front.}
The hidden set is frame-disjoint from the release (minimum gap 301 frames against all 81{,}856 released MorphoCam frames), but not session-disjoint: 48\% of its frames lie within one minute of a released snippet (median 58\,s), and a second, public rig recorded the same enclosures on the same days.
It therefore measures \emph{in-domain} performance, not cross-site transfer.
Its RGB--LiDAR extrinsic is one static calibration per record; we attempted automatic per-scene refinement and rejected it after independent validation (Sec.~\ref{sec:exp_envelope}), and instead rely on the mask-coherence filter of step~(6), which reduced the median in-mask depth IQR from 0.84 to 0.65\,m at the cost of 140 detections.

\subsection{Evaluation Protocols and Metrics}
\label{sec:metrics}

\paragraph{Camera-to-animal distance (T1, T3).}
For each detection, SAM2 segments the animal inside its MegaDetector box; ground truth $d^\ast$ is the median LiDAR depth inside the mask.
A method's prediction $\hat d$ is read from its depth map at the mask centroid.
We report RMSE, AbsRel $=|\hat d-d^\ast|/d^\ast$, SILog, and $\delta{<}1.25$, per distance bin and overall.
We evaluate each MDE in two modes.
\emph{Uncalibrated} (UC): a single scale $s=d^{\text{lidar}}/d^{\text{mde}}$ from the one LiDAR point nearest the image centre.
\emph{Calibrated}: a per-frame affine map $D^{\text{metric}}=sD^{\text{mde}}+t$ fitted by robust least squares to a set of metric anchors.
The anchor set defines the variant: all LiDAR points in the frame (\emph{LC-scene}); LiDAR points inside animal masks (\emph{LC-box}); or object-level radar ranges (\emph{RC}, below).
Because MDEs such as MoGe-2 over-predict far depth, a single scene-wide affine fit is biased by distant background; we restrict anchors to $<$55\,m.

\paragraph{Dense depth (T2).}
Per-pixel RMSE, MAE, AbsRel and $\delta{<}1.25$ against the 11-sweep accumulated LiDAR, reported separately on animal-mask pixels and background pixels, together with coverage (fraction of test frames submitted) so that partial submissions cannot win by cherry-picking.

\paragraph{Motion segmentation (T4).}
Precision is the fraction of points labelled \emph{moving} that fall inside an animal box; densification is the ratio of moving LiDAR points to moving radar returns. Agreement with an independent motion label is measured per animal against Farneb\"ack optical flow~\cite{farneback2003two} between consecutive frames (median flow in the mask minus background median flow $>$1\,px).

\paragraph{Metric 3D tracking (T5).}
Given 2D tracks, each track is lifted to a metric 3D trajectory $\{\mathbf p_k\}$ by back-projecting the mask centroid at a depth given by LiDAR, radar, or calibrated MDE.
Against LiDAR-lifted reference trajectories we report 3D position RMSE, distance AbsRel, speed error (on tracks whose reference speed exceeds 0.3\,m/s), and the frame-to-frame jitter of the raw distance. For distance sampling we fit a half-normal detection function to per-track distances and report the bias of its scale $\sigma$ and a two-sample KS test against LiDAR. Association quality (HOTA~\cite{luiten2021hota}, IDF1/MOTA~\cite{bernardin2008clearmot}) requires identity labels and is left to future work.

\subsection{Baseline Methods}
\label{sec:methods}

\paragraph{Radar-anchored metric scale (RC).}
Radar returns are transformed to the LiDAR frame and then to the camera.
For each animal box $b$ we collect the radar returns whose projection falls within the box columns (azimuth is accurate) and within the box rows padded by one box height on either side (elevation is not).
These returns mix the animal with ground and background seen past it, so their median is biased far.
We compare four LiDAR-free estimators of the animal's range $r_b$ -- the median, the nearest cluster of $\geq$3 returns within 2\,m, the densest such cluster, and the median of the returns within $\pm$35\% of the MDE's own metric prediction, and select one on development sequences (Sec.~\ref{sec:exp_radar_scale}).
The anchor pairs $\{(D^{\text{mde}}_b, r_b)\}$, with $D^{\text{mde}}_b$ the robust median MDE depth in the animal mask, are used to fit the per-frame map: affine with two or more anchors, scale-only with one.
Associating at the object level is what makes radar usable at all: per-pixel association inherits the 1--2\,m elevation error of the extrinsic.

\paragraph{Doppler transfer to LiDAR (motion masks).}
Radar marks motion but sparsely; LiDAR is dense but static in time.
We transfer Doppler to LiDAR by nearest-neighbour search in an \emph{anisotropic} metric: each LiDAR point inherits the Doppler of the nearest radar return within an ellipsoid of radii $(0.6, 1.0, 1.5)$\,m in $(x,y,z)$, the elongated vertical axis absorbing the radar's elevation error.
LiDAR points inheriting $|h|>0.5$\,m/s are labelled moving.
Doppler on a moving platform is dominated by ego-motion, which follows a cosine in azimuth; per frame we fit $h(\phi)=v_x\cos\phi+v_y\sin\phi+c$ with iterative inlier re-fitting~\cite{kellner2013instantaneous} and treat the frame as ego-motion if $R^2>0.5$ or the fitted speed exceeds 1\,m/s with more than 70\% inliers.

\paragraph{Motion-aware LiDAR accumulation.}
A single LiDAR sweep is too sparse on a distant animal, so dense depth is obtained by accumulating sweeps over a short window and fusing them with an MDE prior (MoGe-2) via RANSAC scale-shift alignment on SAM2-masked points, with a Sobel boundary filter against telephoto edge artefacts~\cite{fischlerRandomSampleConsensus1981,birkl2023midas,kanopoulosDesignImageEdge1988}.
Accumulation is valid only for still animals; a moving animal is smeared.
Our gated variant marks an animal as moving when at least three of the LiDAR points in its mask inherit $|h|>0.5$\,m/s through the Doppler transfer and, for those animals only, uses the current sweep alone. We evaluate on static-platform frames, where past sweeps are already registered, with windows of 5, 15 and 30 sweeps, and compare against a gate driven by optical flow.

\paragraph{Metric track lifting.}
2D tracks are formed by SORT-style IoU association~\cite{bewley2016sort} of MegaDetector detections with Hungarian matching (IoU $\geq$0.3, tracks kept alive for 5 frames).
Each observation is back-projected at its mask centroid with a depth from one of the sources -- LiDAR (reference), MoGe-2, UC or LC-scene per frame, a LiDAR scale fixed on the first frame of the track, object-level radar range, or MoGe-2 scaled by the median radar/MoGe-2 ratio over the track -- and every trajectory is smoothed with the same constant-velocity Kalman filter and RTS smoother~\cite{kalman1960new}.

%% file: sections/5_experiments.tex
\section{Experiments}
\label{sec:experiments}

We first re-establish the RGB--LiDAR baselines on the released data and the hidden test (Secs.~\ref{sec:exp_mde}--\ref{sec:exp_trackb}), then study what radar adds for metric scale (Sec.~\ref{sec:exp_radar_scale}) and motion (Secs.~\ref{sec:exp_doppler}--\ref{sec:exp_motion_dens}), evaluate metric 3D tracking (Sec.~\ref{sec:exp_tracking}), and close with radar's operating envelope (Sec.~\ref{sec:exp_envelope}).
LiDAR is the metric reference throughout.
All radar-subset experiments run on every synchronized July-8 frame (19{,}716 frames): MegaDetector~V6 (confidence $\geq$0.3), SAM2.1 masks, and MoGe-2 depth are computed once per frame, and every number is produced by a released script (\emph{Code availability}).
Ground truth uses the anchor LiDAR$\rightarrow$camera extrinsic of each sequence (Sec.~\ref{sec:radar_data}).

\subsection{Monocular Distance to Animals (T1)}
\label{sec:exp_mde}

\begin{figure*}[t]
\centering
\begin{subfigure}{0.48\textwidth}
\figph{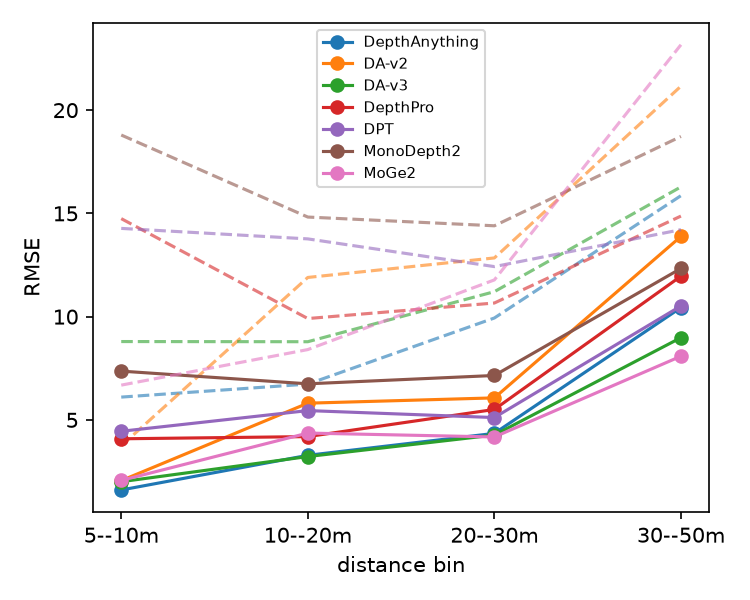}{\linewidth}{RMSE vs distance}
\caption{RMSE vs.\ distance}
\end{subfigure}\hfill
\begin{subfigure}{0.48\textwidth}
\figph{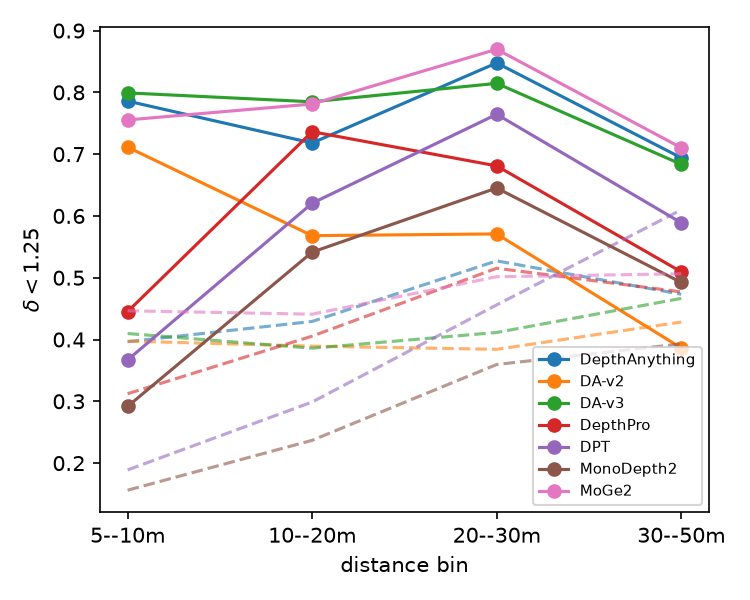}{\linewidth}{delta vs distance}
\caption{$\delta{<}1.25$ vs.\ distance}
\end{subfigure}
\caption{Per-distance camera-to-animal error on the released snippets. Solid: LiDAR-calibrated; dashed: uncalibrated.}
\label{fig:per_distance_metrics}
\end{figure*}

\paragraph{Released snippets.}
We evaluate seven MDEs -- Depth Anything v1/v2/v3~\cite{yang2024depth}, DPT~\cite{ranftl2021vision}, Depth Pro~\cite{bochkovskii2024depth}, MonoDepth2~\cite{li2024improved}, and MoGe-2~\cite{wangMoGe2AccurateMonocular2025} -- on 11{,}498 animal detections from 3{,}695 frames of the released snippets (every 10th frame, at most 40 per snippet; re-weighting by true frame counts moves the median ground-truth distance only from 20.5 to 21.3\,m).
Table~\ref{tab:mde_overall} summarizes the results.
Uncalibrated predictions are unreliable in absolute terms (AbsRel 0.38--0.83, $\delta{<}1.25$ of 0.26--0.46).
Per-frame LiDAR calibration roughly halves RMSE for every model and raises $\delta{<}1.25$ to 0.50--0.79.
MoGe-2, Depth Anything v3 and v1 are the strongest calibrated models (AbsRel $\approx$0.15), and error is highest in the 30--50\,m bin for every model (Fig.~\ref{fig:per_distance_metrics}).

\begin{table}[t]
\centering
\caption{Camera-to-animal distance on the released snippets, all bins (11{,}498 detections, 5--50\,m). UC = uncalibrated; LC = per-frame LiDAR-calibrated (scene anchors).}
\label{tab:mde_overall}
\resizebox{\columnwidth}{!}{%
\begin{tabular}{l|cc|cc|cc|cc}
\toprule
 & \multicolumn{2}{c|}{RMSE (m)$\downarrow$} & \multicolumn{2}{c|}{AbsRel$\downarrow$} & \multicolumn{2}{c|}{SILog$\downarrow$} & \multicolumn{2}{c}{$\delta{<}1.25\uparrow$} \\
Model & UC & LC & UC & LC & UC & LC & UC & LC \\
\midrule
DepthAnything & 8.69 & 4.43 & 0.38 & 0.155 & 0.72 & 0.245 & 0.45 & 0.767 \\
DA-v2         & 11.91 & 6.42 & 0.41 & 0.210 & 0.99 & 0.274 & 0.39 & 0.590 \\
DA-v3         & 10.33 & \textbf{4.14} & 0.47 & 0.153 & 0.82 & 0.225 & 0.41 & 0.787 \\
DepthPro      & 12.03 & 5.66 & 0.56 & 0.243 & 0.82 & 0.323 & 0.42 & 0.625 \\
DPT           & 13.62 & 5.79 & 0.72 & 0.278 & 0.70 & 0.293 & 0.34 & 0.588 \\
MonoDepth2    & 16.23 & 7.68 & 0.83 & 0.381 & 0.98 & 0.402 & 0.26 & 0.498 \\
MoGe-2        & 11.02 & 4.36 & 0.41 & \textbf{0.152} & 0.81 & \textbf{0.219} & 0.46 & \textbf{0.791} \\
\bottomrule
\end{tabular}}
\end{table}

\paragraph{Hidden test, Track~A.}
Table~\ref{tab:trackA} reports organiser baselines on the frozen hidden test, scored by the public scoring program against the checksum-verified reference.
MoGe-2 is clearly strongest (LC AbsRel 0.094), followed by the metric-outdoor Depth Anything V2 (0.169) and the diffusion-based Marigold (0.252).
Models that predict relative \emph{disparity} (DPT, relative Depth Anything V2) are handicapped when inverted to depth for an RGB-only submission, because the scorer's affine alignment operates in depth; we report them for completeness but recommend metric or affine-invariant-depth models for Track~A.

\begin{table}[t]
\centering
\caption{Hidden test, Track~A (RGB only; 510 detections in 352 frames, 13 scenes), scored by the competition's scoring program. $^\dagger$Relative disparity inverted to depth. $^\ddagger$Affine-invariant depth (UC not meaningful).}
\label{tab:trackA}
\resizebox{\columnwidth}{!}{\input{tables/tab_e1_trackA.tex}}
\end{table}

\subsection{Depth Completion from One LiDAR Sweep (T2)}
\label{sec:exp_trackb}

Track~B asks for dense depth from RGB and one raw LiDAR sweep.
Table~\ref{tab:trackB} shows that the split into animal and background pixels matters: a nearest-neighbour fill of the sweep is 35\% worse on animals (RMSE 6.10\,m) than on background (4.51\,m), and aligning MoGe-2 to the sweep improves the animals (5.33\,m) while leaving the background unchanged.
Neither baseline closes the animal gap, which is the headroom the track is designed to measure.

\begin{table}[t]
\centering
\caption{Hidden test, Track~B (RGB + one LiDAR sweep; 902 frames). Per-pixel error on animal-mask and background pixels.}
\label{tab:trackB}
\resizebox{\columnwidth}{!}{\input{tables/tab_e1_trackB.tex}}
\end{table}

\subsection{Radar as a Metric Anchor (T3)}
\label{sec:exp_radar_scale}

Can object-level radar range replace LiDAR as the anchor that gives monocular depth its metric scale?
We compare anchor sources on 48{,}723 animals in 18{,}879 frames (Table~\ref{tab:radar_scale}).
All methods are scored against the same held-out ground truth: each mask's LiDAR points are split at random, one half defines the ground truth and the other half is the LC-box anchor, so no LiDAR baseline is scored on its own anchor points.
The LiDAR-free radar range estimator was chosen on odd-numbered (development) sequences (Table~\ref{tab:estimator}) and all results are reported on the even-numbered (test) sequences.

\begin{table*}[t]
\centering
\caption{Metric calibration of MoGe-2 by anchor source: camera-to-animal distance on the test sequences, frames that contain at least one radar-associated animal (20{,}869 animals). Sensor used at inference: L = LiDAR, R = radar. 95\% CIs from a frame-level bootstrap. $^\ast$The anchor used in an earlier version of this study, which snaps radar returns to their nearest LiDAR point and uses the LiDAR depth.}
\label{tab:radar_scale}
\input{tables/tab_e3_radar_scale.tex}
\end{table*}

\begin{table}[t]
\centering
\caption{LiDAR-free radar range estimators, development sequences (errors relative to held-out LiDAR).}
\label{tab:estimator}
\resizebox{0.6\columnwidth}{!}{\input{tables/tab_e3_estimator_selection.tex}}
\end{table}

\begin{figure}[t]
\centering
\figph{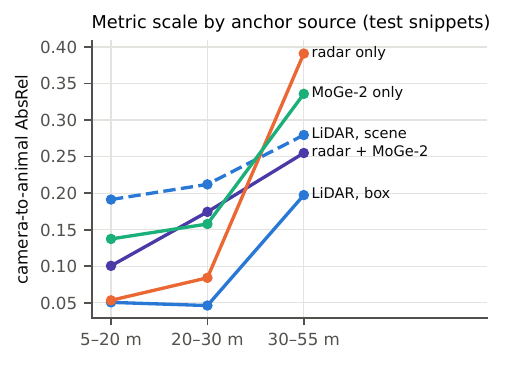}{0.6\columnwidth}{AbsRel per distance bin by anchor}
\caption{Camera-to-animal AbsRel by distance and anchor source (test sequences). Radar range is the best LiDAR-free cue below 30\,m and the worst beyond it, where few returns land on the animal.}
\label{fig:radar_scale_range}
\end{figure}

Four findings follow.
(i)~\emph{Radar is usable only at the object level.} Per-pixel radar anchors (RC-pixel) are worse than no anchor at all (AbsRel 0.48), because the elevation error places returns on the wrong image rows.
(ii)~\emph{Object-level radar beats scene-level LiDAR calibration and the MDE alone, without LiDAR.} RC-box reaches AbsRel 0.120 (median error 0.60\,m) against 0.198 for the scene-level LiDAR calibration used in WildDepth and 0.146 for MoGe-2's own metric output.
A scene-wide affine fit is dominated by ground and background points, which is why it can hurt a model that is already metric.
(iii)~\emph{Where radar sees the animal, its range is nearly as good as LiDAR.} On the 15{,}196 test animals with an associated radar range (73\%), that range alone has AbsRel 0.064, against 0.144 for MoGe-2 and 0.046 for held-out LiDAR on the same animals.
The best estimator gates the returns in the box with MoGe-2's metric prediction ($\pm$35\%) and is therefore an RGB+radar method; a radar-only estimator (nearest range cluster) reaches a 4.2\% median relative error at 90\% coverage but has a heavy tail.
(iv)~\emph{Radar does not replace LiDAR.} Box-level LiDAR remains more than twice as accurate (0.052 vs.\ 0.120), radar's advantage vanishes beyond 30\,m (Fig.~\ref{fig:radar_scale_range}), and animals without their own radar anchor inherit a frame-level fit that is much less accurate (leave-own-anchor-out AbsRel 0.239).
We also note a correction to an earlier analysis of these data: the anchor that snaps radar to LiDAR (last row) needs LiDAR at inference and does not replicate at scale (AbsRel 0.572).
The conclusions are unchanged when the per-frame extrinsic is used for ground truth instead of the anchor extrinsic.

\subsection{Doppler as a Motion Cue (T4)}
\label{sec:exp_doppler}

\paragraph{Where is Doppler on animals?}
On the vehicle-mounted rig most Doppler is ego-motion: the per-frame cosine fit flags 35\% of frames as ego-motion frames, which we exclude, leaving 12{,}596 static-platform frames that contain animals.
In these, Doppler above 0.5\,m/s is associated with 8.5\% of animal detections, concentrated in walking herds.

\begin{figure*}[t]
\centering
\figph{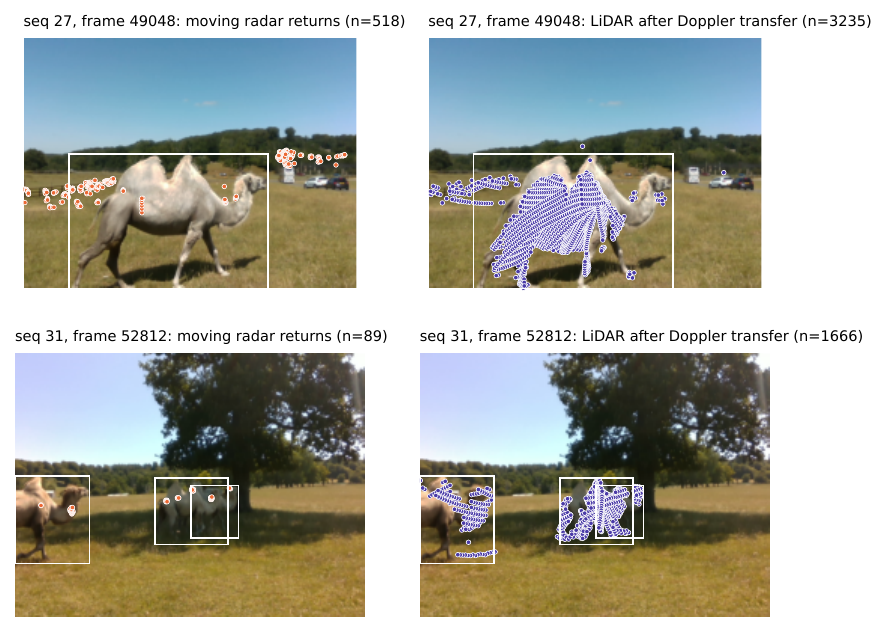}{0.9\textwidth}{Doppler transfer examples}
\caption{Doppler transfer. Sparse moving radar returns (left, projected with their elevation error) become a dense motion mask on the co-registered LiDAR (right) for walking camels.}
\label{fig:transfer}
\end{figure*}

\paragraph{Doppler transfer yields dense motion masks.}
In the 2{,}783 static-platform frames where an animal box contains at least three moving radar returns, transferring Doppler to single-sweep LiDAR (Sec.~\ref{sec:methods}) multiplies the number of moving points by 3.7$\times$; 72\% of transferred moving LiDAR points fall on an animal, against 65\% of the raw moving radar returns (Table~\ref{tab:transfer}).
Precision varies by sequence from 0.39 -- where moving safari vehicles share the view -- to 0.96--0.98 on walking camel herds (Fig.~\ref{fig:transfer}).
The anisotropic association radius is what makes the transfer dense: an isotropic 0.6\,m radius yields only 1.75$\times$ densification at the same precision.
Doppler and optical flow flag different animals (precision 0.57, recall 0.28 of radar against a flow motion label), as expected since radar measures radial and flow lateral motion.

\begin{table}[t]
\centering
\caption{Doppler transfer from radar to single-sweep LiDAR (static-platform event frames). Densification = moving LiDAR points / moving radar returns; precision = fraction inside animal boxes.}
\label{tab:transfer}
\resizebox{\columnwidth}{!}{\input{tables/tab_e4_transfer_ablation.tex}}
\end{table}

\subsection{Motion-Aware Densification}
\label{sec:exp_motion_dens}

Temporal LiDAR accumulation gives dense, coherent surfaces for still animals (Fig.~\ref{fig:examples_tpd}) but can smear moving ones.
We test whether Doppler gating (Sec.~\ref{sec:methods}) helps, with identical LiDAR densification in all rows and per-pixel error measured against a held-out half of the current sweep (Table~\ref{tab:radar_dens}).
The answer depends on the accumulation window.
Over 0.5\,s (5 sweeps), accumulation helps even moving animals and gating is unnecessary.
Over 1.5\,s gating reduces the error on Doppler-flagged animals by 11\%, and over 3\,s by 37\% (2.60$\rightarrow$1.64\,m), with gate precision rising from 0.54 to 0.85.
Across all animals at 3\,s, an optical-flow gate is better than the Doppler gate (2.08 vs.\ 2.17\,m) because it fires on more of the movers; Doppler is the more precise gate, and the only one that does not depend on illumination.

\begin{table}[t]
\centering
\caption{Doppler-gated LiDAR accumulation: dense-depth MAE (m) against held-out single-sweep LiDAR, by accumulation window. Gate precision = fraction of gated animals whose error decreases.}
\label{tab:radar_dens}
\resizebox{\columnwidth}{!}{\input{tables/tab_e5_gated_accumulation.tex}}
\end{table}

\subsection{Metric 3D Animal Tracking (T5)}
\label{sec:exp_tracking}

We now use these depth sources for the task a tracker must solve: turning 2D animal tracks into metric trajectories.
An IoU tracker links detections over all 19{,}716 frames; the 431 tracks with at least 20 observations (50{,}338 observations) are lifted to 3D with each depth source and smoothed with the same constant-velocity Kalman/RTS smoother, and the LiDAR-lifted track is the reference (Table~\ref{tab:track3d}).

\begin{table*}[t]
\centering
\caption{Metric 3D tracking on the radar subset (431 tracks). Errors against LiDAR-lifted trajectories (median over tracks); speed error on 169 tracks with reference speed $>$0.3\,m/s; jitter = std of the raw frame-to-frame distance change; $\sigma$ bias = bias of the half-normal detection-function scale fitted to per-track distances (distance sampling).}
\label{tab:track3d}
\input{tables/tab_e6_tracking.tex}
\end{table*}

\paragraph{Trajectory accuracy.}
MoGe-2 alone places animals 2.2\,m from their true position (distance AbsRel 0.149), and per-frame calibration from a single LiDAR point (UC) is far worse (6.3\,m), because its frame-to-frame scale jitter (1.7\,m) becomes spurious motion.
Per-frame scene-level LiDAR calibration reduces the error to 1.5\,m.
Calibrating once per \emph{track} is much better than calibrating every frame: a LiDAR scale taken from the first frame only gives 0.52\,m.
Without any LiDAR, scaling MoGe-2 by the median radar/MoGe-2 ratio over the track gives 0.58\,m position error, 3.3\% distance error and 0.10\,m/s speed error, matching the once-per-track LiDAR calibration and improving on MoGe-2 alone by a factor of four.
Radar range alone is equally accurate in position but twice as jittery (0.72\,m).

\begin{figure}[t]
\centering
\figph{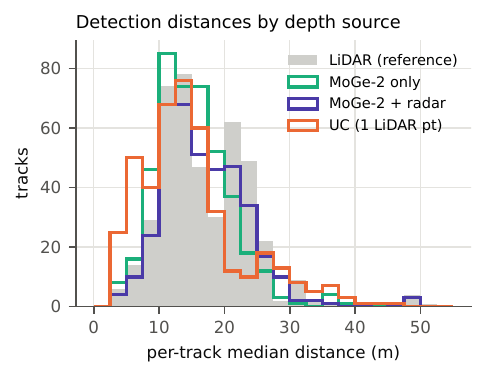}{0.7\columnwidth}{per-track distance histograms}
\caption{Per-track detection distances from LiDAR (grey), MoGe-2 alone, MoGe-2 scaled by radar per track, and single-point LiDAR calibration (UC).}
\label{fig:distsampling}
\end{figure}

\paragraph{Application: distance sampling.}
Distance-sampling abundance estimates depend on the \emph{distribution} of detection distances~\cite{buckland2001distance,howe2017distance}.
Fitting a half-normal detection function to per-track distances (Fig.~\ref{fig:distsampling}), MoGe-2 alone underestimates its scale $\sigma$ by 13\% (the distance distribution differs from LiDAR's, KS $p<10^{-6}$); with a per-track radar scale the bias falls to $-$2.1\% and the distributions are statistically indistinguishable (KS $p=0.58$).

\begin{figure}[t]
\centering
\figph{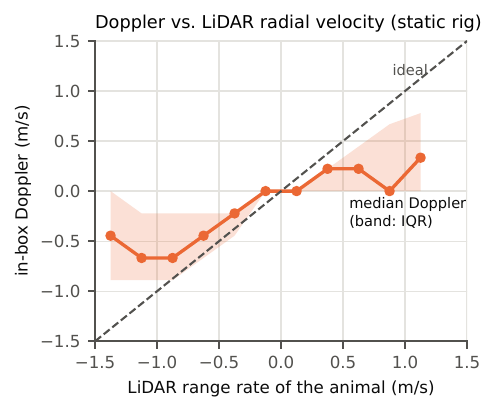}{0.6\columnwidth}{binned Doppler vs LiDAR range rate}
\caption{In-box Doppler against LiDAR-derived range rate (static platform; median and inter-quartile band per bin).}
\label{fig:doppler_vel}
\end{figure}

\paragraph{Can Doppler measure animal speed?}
For 29{,}347 track observations with radar returns at the animal's range, in-box Doppler correlates with the LiDAR range rate at $r=0.57$ ($r=0.61$ on the static platform).
On moving animals (range rate above 0.3\,m/s, $n=3{,}080$) the correlation is $r=0.60$ and the sign agrees 72\% of the time, but Doppler under-reads the radial speed by about half (Theil--Sen slope 0.47, median error 0.29\,m/s; Fig.~\ref{fig:doppler_vel}), consistent with limb micro-Doppler~\cite{chen2006microdoppler} and static returns mixing into the box.
Doppler thus measures the direction and approximate magnitude of radial motion; it is a usable, if biased, velocity observation for a tracker.
The spread of in-box Doppler also increases with speed (Spearman 0.54), which suggests it as a cue for activity.
This corrects an earlier analysis of these data that reported no correlation.

\subsection{Radar's Operating Envelope for Wildlife}
\label{sec:exp_envelope}

\begin{figure*}[t]
\centering
\figph{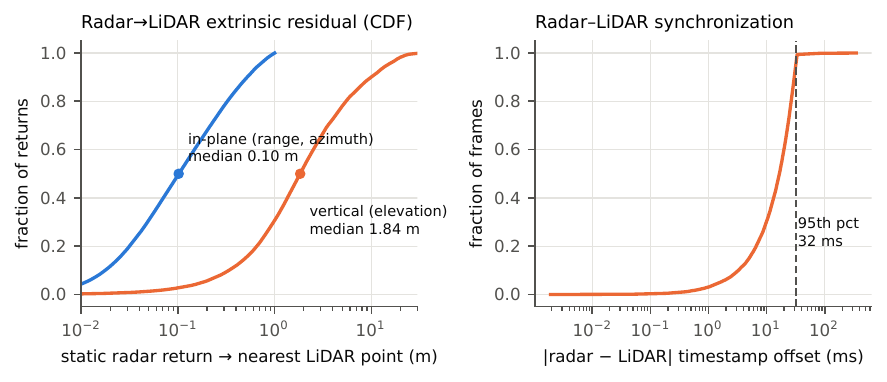}{0.85\textwidth}{calibration budget}
\caption{Radar error budget. Left: residual of static radar returns to the nearest LiDAR point, in the ground plane (range/azimuth; matches gated at 1\,m) and vertically. Right: radar--LiDAR timestamp offset.}
\label{fig:calib}
\end{figure*}

\noindent\textbf{Elevation error dominates.}
Over 827k static radar returns matched to LiDAR, the radar extrinsic residual has a median of 0.10\,m in the ground plane but 1.84\,m vertically (Fig.~\ref{fig:calib}); this is why per-pixel radar anchoring fails and every positive result above uses object-level association.

\noindent\textbf{No coverage extension and no foliage penetration.}
Moving radar returns without nearby LiDAR exist but are vehicles, foliage, or far clutter -- none are animals after a depth-consistency check -- and a search for animals seen through vegetation found only artefacts; the Longleat enclosures are open grassland, and the vegetated sites have no radar.

\noindent\textbf{Doppler does not clean calibration anchors.}
Removing moving points from the anchors changes the calibration error by +0.01\,m over 2{,}426 animals in an earlier analysis on 32 sequences.

\noindent\textbf{Automatic RGB--LiDAR offset refinement fails.}
For the MorphoCam rig, maximizing agreement between LiDAR depth edges and RGB gradients over image shifts produced optima that tracked the search bound, and 11 of 13 scenes failed an independent in-mask spread check; edge agreement has no interior optimum on sparse LiDAR in natural scenes.

\noindent\textbf{Per-frame extrinsic refinement can drift.}
The per-frame LiDAR$\rightarrow$camera extrinsics of the CPS2 rig drift away from the sequence anchor by up to 25$^\circ$; where the drift exceeds 5$^\circ$, the anchor extrinsic puts more than twice as many LiDAR points on the animal (116 vs.\ 50) with a tighter in-mask depth spread (0.83 vs.\ 1.27\,m).
We therefore use the anchor extrinsic throughout (Sec.~\ref{sec:radar_data}).

\subsection{Densification, Behaviour and 4D Reconstruction (from WildDepth)}
\label{sec:exp_eccv_tasks}

\begin{figure}[t]
\centering
\figph{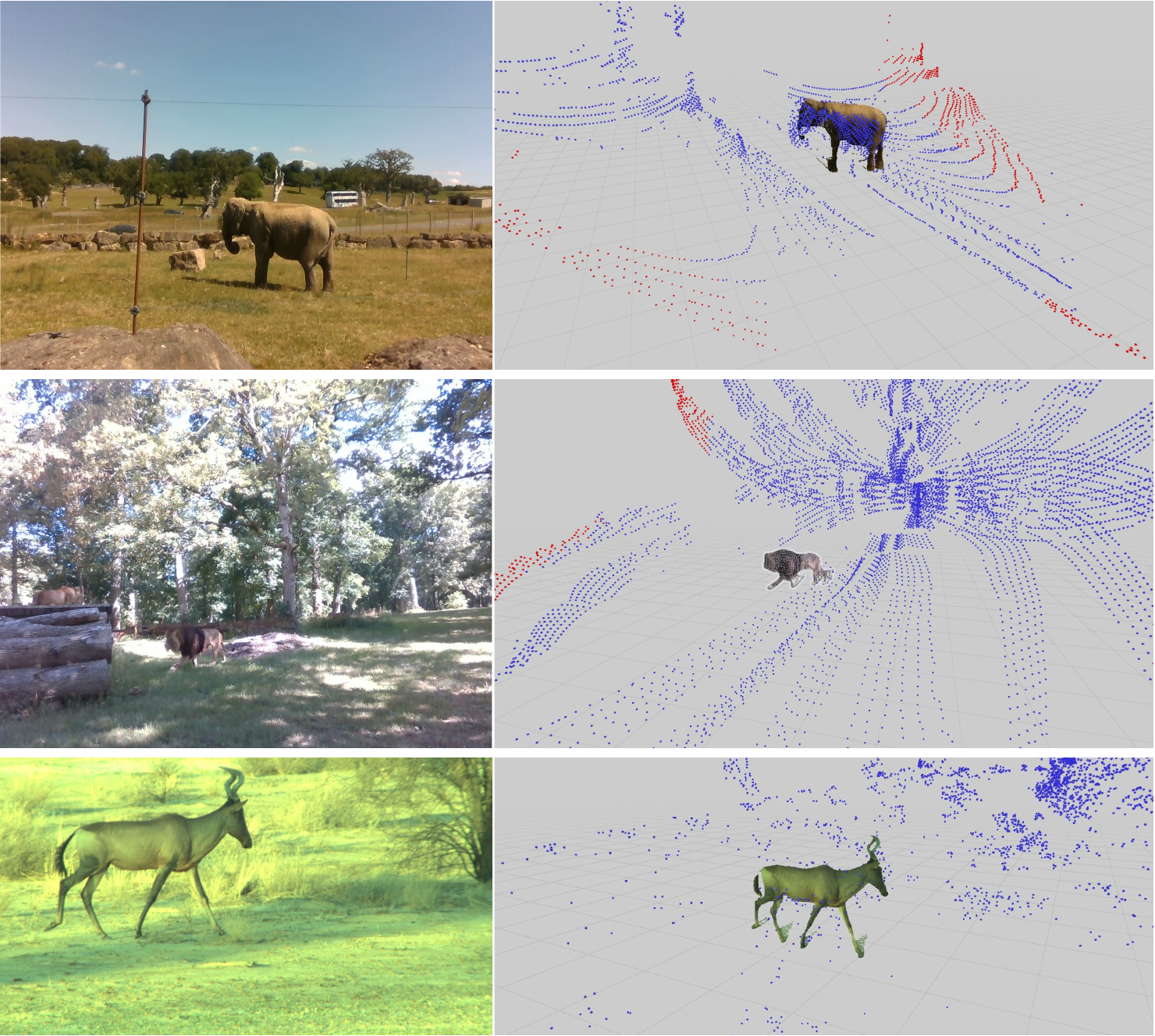}{\columnwidth}{target point densification}
\caption{Target-point densification: RGB (left) and dense metric point cloud (right) for an elephant and a lion (Longleat) and a red hartebeest (Kgalagadi).}
\label{fig:examples_tpd}
\end{figure}

\begin{figure*}[t]
\centering
\figph{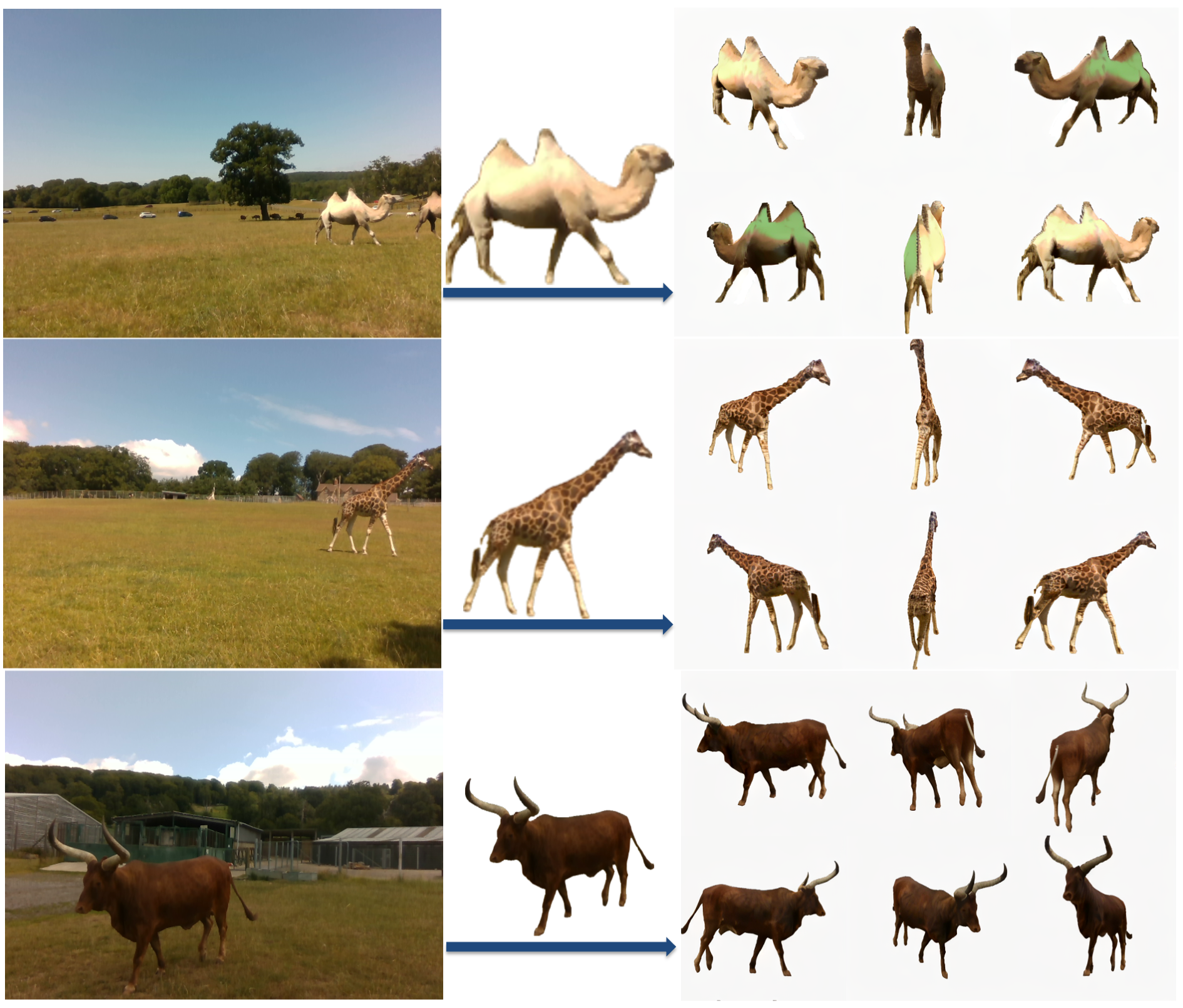}{\textwidth}{4D reconstructions}
\caption{4D reconstructions of a walking camel (top), giraffe (middle) and Ankole cow (bottom).}
\label{fig:examples_3DR}
\end{figure*}

For completeness we summarize the remaining WildDepth baselines~\cite{aamirWildDepth2026}.
\emph{Densification:} on a walking red hartebeest at ${\approx}$97\,m (Kgalagadi), MoGe-2 + LiDAR scale-shift alignment reaches positive $R^2$ (0.25) where a planar fit does not ($-$0.10), at similar MAE (Table~\ref{tab:ablation_results}, Fig.~\ref{fig:examples_tpd}).
\emph{Behaviour:} VideoMAE~\cite{wang2023videomae} classifiers over 12 activities and 20 fine-grained actions (per-class F1 in the supplementary material).
\emph{4D reconstruction:} SAM2 masks with a Gaussian-splatting deformation field~\cite{feng2025gaussian} (Fig.~\ref{fig:examples_3DR}).
\emph{Robustness to the MDE:} repeating the anchor comparison with the diffusion-based Marigold~\cite{ke2024marigold} on a 233-frame subset leaves the conclusions unchanged (supplementary Table~S1): with anchors, Marigold is on par with MoGe-2 (e.g.\ RC-box 0.131 vs.\ 0.137), but it is affine-invariant and cannot be used without an anchor.

\begin{table}[t]
\centering
\caption{RGB--LiDAR densification, red hartebeest at ${\approx}$97\,m.}
\label{tab:ablation_results}
\begin{tabular}{lrr}
\toprule
Method & MAE (m)$\downarrow$ & $R^2\uparrow$ \\
\midrule
Planar depth & $0.13 \pm 1.14$ & $-0.10 \pm 0.13$ \\
MoGe-2 + SSA & $0.16 \pm 1.12$ & $0.25 \pm 1.36$ \\
\bottomrule
\end{tabular}
\end{table}

%% file: tables/tab_e1_trackA.tex
\begin{tabular}{lccccc}
\toprule
Model & LC AbsRel & LC RMSE (m) & LC SILog & LC $\delta{<}1.25$ & UC AbsRel \\
\midrule
MoGe-2 & 0.094 & 3.25 & 0.206 & 0.97 & 0.335 \\
Depth Anything V2 (metric outdoor) & 0.169 & 4.16 & 0.168 & 0.89 & 0.458 \\
Marigold v1.1$^\ddagger$ & 0.252 & 6.51 & 0.247 & 0.40 & 0.359 \\
DPT$^\dagger$ & 0.274 & 6.57 & 0.337 & 0.65 & 0.871 \\
Depth Pro & 0.337 & 7.14 & 0.375 & 0.53 & 3.064 \\
Depth Anything V2 (relative)$^\dagger$ & 0.398 & 9.21 & 0.470 & 0.22 & 0.317 \\
\bottomrule
\end{tabular}

%% file: tables/tab_e1_trackB.tex
\begin{tabular}{lccccc}
\toprule
Method & RMSE animal & RMSE bg. & AbsRel animal & AbsRel all & $\delta{<}1.25$ \\
\midrule
Nearest-neighbour fill & 6.10 & 4.51 & 0.191 & 0.122 & 0.92 \\
MoGe-2 + scale-shift to sweep & 5.33 & 4.52 & 0.193 & 0.203 & 0.80 \\
\bottomrule
\end{tabular}

%% file: tables/tab_e3_radar_scale.tex
\begin{tabular}{llcccc}
\toprule
Anchor & Sensor & AbsRel [95\% CI] & medAE (m) & $\delta{<}1.25$ & cov. \\
\midrule
MoGe-2 metric, no anchor & -- & 0.146 [0.145, 0.148] & 1.93 & 0.80 & 1.00 \\
UC: 1 LiDAR point (scale) & L & 0.442 [0.437, 0.448] & 6.53 & 0.21 & 1.00 \\
LC-scene: all LiDAR $<$55\,m & L & 0.198 [0.196, 0.201] & 2.49 & 0.55 & 1.00 \\
LC-box: held-out LiDAR in masks & L & 0.052 [0.050, 0.054] & 0.17 & 0.95 & 1.00 \\
RC-pixel: radar returns per pixel & R & 0.483 [0.471, 0.495] & 4.54 & 0.41 & 1.00 \\
RC-box: object-level radar (affine) & R & 0.120 [0.113, 0.131] & 0.60 & 0.85 & 1.00 \\
RC-box, scale only & R & 0.123 [0.121, 0.125] & 0.85 & 0.81 & 1.00 \\
RC-box, leave-own-anchor-out & R & 0.239 [0.208, 0.277] & 1.28 & 0.75 & 0.87 \\
Radar snapped to LiDAR (earlier study)$^\ast$ & L+R & 0.572 [0.549, 0.602] & 2.04 & 0.58 & 1.00 \\
\midrule
\multicolumn{6}{l}{\emph{Same animals, where a radar range estimate exists (15{,}196 animals)}} \\
MoGe-2 metric, no anchor & -- & 0.144 [0.142, 0.146] & 1.92 & 0.80 & -- \\
LC-box: held-out LiDAR & L & 0.046 [0.043, 0.049] & 0.13 & 0.96 & -- \\
Radar range only (MoGe-gated) & R & 0.064 [0.063, 0.066] & 0.41 & 0.94 & -- \\
\bottomrule
\end{tabular}

%% file: tables/tab_e3_estimator_selection.tex
\begin{tabular}{lccc}
\toprule
Radar range estimator & coverage & median rel.\ err. & AbsRel \\
\midrule
median of box returns & 0.90 & 0.064 & 0.728 \\
nearest range cluster & 0.90 & 0.042 & 0.349 \\
densest range cluster & 0.90 & 0.052 & 0.763 \\
MoGe-2-gated ($\pm$35\%) & 0.76 & 0.026 & 0.056 \\
\bottomrule
\end{tabular}

%% file: tables/tab_e4_transfer_ablation.tex
\begin{tabular}{llcccc}
\toprule
Radii (m) & $\tau$ (m/s) & event frames & densif. & prec.\ radar & prec.\ LiDAR \\
\midrule
0.6/1.0/1.5 (ours) & 0.3 & 3{,}815 & 4.0$\times$ & 0.64 & 0.75 \\
0.6/1.0/1.5 (ours) & 0.5 & 2{,}783 & 3.7$\times$ & 0.65 & 0.72 \\
0.6/1.0/1.5 (ours) & 0.7 & 1{,}995 & 3.2$\times$ & 0.63 & 0.65 \\
0.6/1.0/1.5 (ours) & 1.0 & 1{,}375 & 2.9$\times$ & 0.62 & 0.57 \\
0.6/0.6/0.6 & 0.5 & 2{,}783 & 1.8$\times$ & 0.65 & 0.72 \\
0.6/0.6/0.6 & 1.0 & 1{,}375 & 1.3$\times$ & 0.62 & 0.55 \\
1.0/1.0/1.0 & 0.5 & 2{,}783 & 3.2$\times$ & 0.65 & 0.72 \\
1.0/1.0/1.0 & 1.0 & 1{,}375 & 2.5$\times$ & 0.62 & 0.60 \\
0.6/1.0/2.5 & 0.5 & 2{,}783 & 4.6$\times$ & 0.65 & 0.69 \\
0.6/1.0/2.5 & 1.0 & 1{,}375 & 3.5$\times$ & 0.62 & 0.53 \\
\bottomrule
\end{tabular}

%% file: tables/tab_e5_gated_accumulation.tex
\begin{tabular}{c|ccc|ccc|cc}
\toprule
 & \multicolumn{3}{c|}{Doppler-moving animals} & \multicolumn{3}{c|}{all animals} & \multicolumn{2}{c}{gate (Doppler)} \\
Sweeps & single & accum. & gated & accum. & Doppler & flow & prec. & rate \\
\midrule
5 (0.5\,s) & 2.02 & 1.96 & 2.02 & 2.05 & 2.05 & 2.06 & 0.54 & 0.09 \\
15 (1.5\,s) & 1.81 & 2.04 & 1.81 & 2.13 & 2.11 & 2.09 & 0.72 & 0.08 \\
30 (3.0\,s) & 1.64 & 2.60 & 1.64 & 2.25 & 2.17 & 2.08 & 0.85 & 0.08 \\
\bottomrule
\end{tabular}

%% file: tables/tab_e6_tracking.tex
\begin{tabular}{llccccc}
\toprule
Depth source & LiDAR & Pos.\ RMSE (m) & Dist.\ AbsRel & Speed err.\ (m/s) & Jitter (m) & $\sigma$ bias \\
\midrule
MoGe-2 metric & -- & 2.20 & 0.149 & 0.15 & 0.31 & -13.2\% \\
UC, per frame & L & 6.27 & 0.399 & 0.28 & 1.72 & -8.1\% \\
LC-scene, per frame & L & 1.52 & 0.089 & 0.11 & 0.28 & -3.6\% \\
LiDAR scale, first frame only & L (1 frame) & 0.52 & 0.031 & 0.10 & 0.35 & +0.2\% \\
Radar range (object level) & R & 0.53 & 0.036 & 0.13 & 0.72 & -2.1\% \\
MoGe-2 $\times$ radar scale (per track) & R & 0.58 & 0.033 & 0.10 & 0.35 & -2.1\% \\
LiDAR (reference) & L & 0 & 0.005 & 0 & 0.12 & 0 \\
\bottomrule
\end{tabular}

%% file: sections/6_discussion.tex
\section{Discussion}
\label{sec:discussion}

\paragraph{What radar is for in the wild.}
Our results separate three roles that are often conflated.
As a \emph{dense depth sensor}, radar is not (yet) competitive with LiDAR: it is much sparser on the animal and its vertical extrinsic error rules out per-pixel use. As radar MIMO antenna count increases with technological improvements, point cloud density and angular resolution will increase, making it a viable dense sensor.
As a \emph{scale anchor}, it is genuinely useful: object-level radar range calibrates a monocular model better than scene-level LiDAR calibration, and on the three quarters of animals where it sees them its range is within 6\% of LiDAR -- on a sensor rig that costs and consumes a fraction of a LiDAR.
As a \emph{motion sensor}, it measures something neither commonly available LiDAR nor a single image can: which returns are moving now, and in which radial direction.
Transferred onto LiDAR it gives dense motion masks, and it can decide when multi-second temporal fusion is safe.

\paragraph{Implications for tracking.}
For metric tracking, the temporal structure of depth error matters more than its per-frame average: a scale that jitters from frame to frame appears as motion.
Calibrating once per track rather than once per frame is therefore the single most effective choice -- it reduced position error three-fold with LiDAR (1.52$\rightarrow$0.52\,m) -- and radar makes it possible without LiDAR (0.58\,m).
For ecological inference the consequence is concrete: monocular depth alone biases the scale of a distance-sampling detection function by $-$13\%, which would inflate density estimates, whereas a per-track radar scale reduces the bias to $-$2\%.
Doppler is a biased but informative velocity observation (r\,=\,0.60, slope 0.47 on moving animals), suitable as a radial-velocity measurement in a tracker's filter once the gain is calibrated.

\paragraph{Implications for sensor design.}
Elevation resolution drives the dominant extrinsic error and blocks per-pixel radar--camera fusion; a static or ego-compensated mount matters because ego-motion swamped the Doppler in 35\% of our frames; and extrinsic calibration should be validated against an independent signal -- in our data the per-frame refined extrinsics drifted by up to 25$^\circ$ while their own optimization objective reported success. An approach to increasingly elevation angular resolution is to use a second radar (V) mounted orthogonally to the original radar (H). In addition, it should be noted that the Altos radar is optimized for automotive applications in terms of its modulation parameters and signal processing chain -- a suggested next step would be to capture raw I/Q mmWave radar for custom signal processing.

\paragraph{Limitations.}
Radar is available at one site, on one day, in daylight and in open enclosures; its illumination invariance and its behaviour in vegetation are observed rather than measured here.
The radar results are strongest below 30\,m, where most of our animals are, and radar range becomes unreliable beyond it.
The best radar range estimator uses the monocular model's metric prediction to select returns and is therefore an RGB+radar method; the purely radar estimator has a heavy tail.
The radar$\rightarrow$LiDAR extrinsic is hand-refined on one sequence, and LiDAR ground truth inherits the calibration and synchronization errors we characterize rather than eliminate.
The tracking reference is itself LiDAR-derived and our tracker is a simple IoU tracker; identity-level metrics (HOTA, IDF1) require identity labels we do not yet have.
The hidden test set is in-domain (not session-disjoint), small in Track~A (510 detections), and contains visible visitors and number plates that must be anonymized before public release.

\paragraph{Future work.}
A night-time RGB--thermal--radar collection (the CPS2 rig also carries a thermal camera) is the natural next step to measure the low-light regime that most motivates radar; identity-verified herd sequences would allow testing whether range and Doppler improve association; and a session-disjoint, multi-site hidden test would turn the benchmark into a generalization test.

%% file: sections/7_conclusion.tex
\section{Conclusion}
\label{sec:conclusion}

\wdtwo{} extends WildDepth from a static RGB--LiDAR depth dataset to a benchmark for metric perception of animals over time.
It adds the first synchronized RGB--LiDAR--radar recordings of animals with a measured calibration and timing error budget, a frozen hidden test with a public scorer, and tasks for metric scale, motion segmentation, and 3D tracking.
Monocular depth models remain meters off at wildlife ranges without a metric anchor; radar supplies that anchor cheaply -- enough to track animals in metric 3D and to remove most of the distance-sampling bias of monocular depth without LiDAR -- and, through Doppler, adds a motion signal that LiDAR lacks, while its elevation error and attenuated Doppler set clear limits.
We release the data, calibration, benchmark, and code, and hope they help turn the millions of existing monocular wildlife images and tracks into metric measurements that ecologists can use.

%% file: egbib.bib
@String(CVPR= {IEEE Conf. Comput. Vis. Pattern Recog.})

@String(ECCV= {Eur. Conf. Comput. Vis.})

@String(ICIP = {IEEE Int. Conf. Image Process.})

@String(CVPR  = {CVPR})

@String(ECCV  = {ECCV})

@String(ICIP  = {ICIP})

@inproceedings{gabeff2025mammalps,
  title={MammAlps: A multi-view video behavior monitoring dataset of wild mammals in the Swiss Alps},
  author={Gabeff, Valentin and Qi, Haozhe and Flaherty, Brendan and Sumbul, Gencer and Mathis, Alexander and Tuia, Devis},
  booktitle={Proceedings of the Computer Vision and Pattern Recognition Conference},
  pages={13854--13864},
  year={2025}
}

@article{duporge2025baboonland,
  title={BaboonLand Dataset: Tracking Primates in the Wild and Automating Behaviour Recognition from Drone Videos},
  author={Duporge, Isla and Kholiavchenko, Maksim and Harel, Roi and Wolf, Scott and Rubenstein, Daniel I and Crofoot, Margaret C and Berger-Wolf, Tanya and Lee, Stephen J and Barreau, Julie and Kline, Jenna and others},
  journal={International Journal of Computer Vision},
  pages={1--12},
  year={2025},
  publisher={Springer}
}

@inproceedings{kholiavchenko2024kabr,
  title={Kabr: In-situ dataset for kenyan animal behavior recognition from drone videos},
  author={Kholiavchenko, Maksim and Kline, Jenna and Ramirez, Michelle and Stevens, Sam and Sheets, Alec and Babu, Reshma and Banerji, Namrata and Campolongo, Elizabeth and Thompson, Matthew and Van Tiel, Nina and others},
  booktitle={Proceedings of the IEEE/CVF Winter Conference on Applications of Computer Vision},
  pages={31--40},
  year={2024}
}

@inproceedings{liu2023lote,
  title={LoTE-Animal: A long time-span dataset for endangered animal behavior understanding},
  author={Liu, Dan and Hou, Jin and Huang, Shaoli and Liu, Jing and He, Yuxin and Zheng, Bochuan and Ning, Jifeng and Zhang, Jingdong},
  booktitle={Proceedings of the IEEE/CVF international conference on computer vision},
  pages={20064--20075},
  year={2023}
}

@inproceedings{yang2024depth,
  title={Depth anything: Unleashing the power of large-scale unlabeled data},
  author={Yang, Lihe and Kang, Bingyi and Huang, Zilong and Xu, Xiaogang and Feng, Jiashi and Zhao, Hengshuang},
  booktitle={Proceedings of the IEEE/CVF conference on computer vision and pattern recognition},
  pages={10371--10381},
  year={2024}
}

@article{bochkovskii2024depth,
  title={Depth pro: Sharp monocular metric depth in less than a second},
  author={Bochkovskii, Aleksei and Delaunoy, Ama{\~A}{\c{G}}l and Germain, Hugo and Santos, Marcel and Zhou, Yichao and Richter, Stephan R and Koltun, Vladlen},
  journal={arXiv preprint arXiv:2410.02073},
  year={2024}
}

@article{birkl2023midas,
  title={Midas v3. 1--a model zoo for robust monocular relative depth estimation},
  author={Birkl, Reiner and Wofk, Diana and M{\"u}ller, Matthias},
  journal={arXiv preprint arXiv:2307.14460},
  year={2023}
}

@article{li2024improved,
  title={An Improved MonoDepth2 Algorithm for Vehicle Monocular Depth Estimation},
  author={Li, Chaoqun and Yue, Chenxi and Liu, Yanyan and Bie, Minglin and Li, Guoning and Lv, Zengming and Li, Jin},
  journal={Optik},
  volume={311},
  pages={171936},
  year={2024},
  publisher={Elsevier}
}

@inproceedings{ranftl2021vision,
  title={Vision transformers for dense prediction},
  author={Ranftl, Ren{\'e} and Bochkovskiy, Alexey and Koltun, Vladlen},
  booktitle={Proceedings of the IEEE/CVF international conference on computer vision},
  pages={12179--12188},
  year={2021}
}

@article{muramatsuWildPoseLongRange3D2024,
  title = {{{WildPose}}: A Long-Range {{3D}} Wildlife Motion Capture System},
  shorttitle = {{{WildPose}}},
  author = {Muramatsu, Naoya and Shin, Sangyun and Deng, Qianyi and Markham, Andrew and Patel, Amir},
  year = 2025,
  month = mar,
  journal = {Journal of Experimental Biology},
  volume = {228},
  number = {5},
  pages = {JEB249987},
  issn = {0022-0949},
  doi = {10.1242/jeb.249987},
  urldate = {2026-05-11}
}

@inproceedings{ng2022animal,
  title={Animal kingdom: A large and diverse dataset for animal behavior understanding},
  author={Ng, Xun Long and Ong, Kian Eng and Zheng, Qichen and Ni, Yun and Yeo, Si Yong and Liu, Jun},
  booktitle={Proceedings of the IEEE/CVF conference on computer vision and pattern recognition},
  pages={19023--19034},
  year={2022}
}

@inproceedings{chen2023mammalnet,
  title={Mammalnet: A large-scale video benchmark for mammal recognition and behavior understanding},
  author={Chen, Jun and Hu, Ming and Coker, Darren J and Berumen, Michael L and Costelloe, Blair and Beery, Sara and Rohrbach, Anna and Elhoseiny, Mohamed},
  booktitle={Proceedings of the IEEE/CVF conference on computer vision and pattern recognition},
  pages={13052--13061},
  year={2023}
}

@article{ma2023chimpact,
  title={Chimpact: A longitudinal dataset for understanding chimpanzee behaviors},
  author={Ma, Xiaoxuan and Kaufhold, Stephan and Su, Jiajun and Zhu, Wentao and Terwilliger, Jack and Meza, Andres and Zhu, Yixin and Rossano, Federico and Wang, Yizhou},
  journal={Advances in Neural Information Processing Systems},
  volume={36},
  pages={27501--27531},
  year={2023}
}

@article{rogers2023meerkat,
  title={Meerkat behaviour recognition dataset},
  author={Rogers, Mitchell and Gendron, Ga{\"e}l and Valdez, David Arturo Soriano and Azhar, Mihailo and Chen, Yang and Heidari, Shahrokh and Perelini, Caleb and O'Leary, Padriac and Knowles, Kobe and Tait, Izak and others},
  journal={arXiv preprint arXiv:2306.11326},
  year={2023}
}

@misc{wangMoGe2AccurateMonocular2025,
  title = {{{MoGe-2}}: {{Accurate Monocular Geometry}} with {{Metric Scale}} and {{Sharp Details}}},
  shorttitle = {{{MoGe-2}}},
  author = {Wang, Ruicheng and Xu, Sicheng and Dong, Yue and Deng, Yu and Xiang, Jianfeng and Lv, Zelong and Sun, Guangzhong and Tong, Xin and Yang, Jiaolong},
  year = 2025,
  month = jul,
  number = {arXiv:2507.02546},
  eprint = {2507.02546},
  primaryclass = {cs},
  publisher = {arXiv},
  doi = {10.48550/arXiv.2507.02546},
  urldate = {2025-07-06},
  archiveprefix = {arXiv}
}

@misc{raviSAM2Segment2024,
  title = {{{SAM}} 2: {{Segment Anything}} in {{Images}} and {{Videos}}},
  shorttitle = {{{SAM}} 2},
  author = {Ravi, Nikhila and Gabeur, Valentin and Hu, Yuan-Ting and Hu, Ronghang and Ryali, Chaitanya and Ma, Tengyu and Khedr, Haitham and R{\"a}dle, Roman and Rolland, Chloe and Gustafson, Laura and Mintun, Eric and Pan, Junting and Alwala, Kalyan Vasudev and Carion, Nicolas and Wu, Chao-Yuan and Girshick, Ross and Doll{\'a}r, Piotr and Feichtenhofer, Christoph},
  year = 2024,
  month = aug,
  number = {arXiv:2408.00714},
  eprint = {2408.00714},
  primaryclass = {cs},
  publisher = {arXiv},
  doi = {10.48550/arXiv.2408.00714},
  urldate = {2024-08-19},
  archiveprefix = {arXiv},
}

@inproceedings{feng2025gaussian,
  title={Gaussian splashing: Unified particles for versatile motion synthesis and rendering},
  author={Feng, Yutao and Feng, Xiang and Shang, Yintong and Jiang, Ying and Yu, Chang and Zong, Zeshun and Shao, Tianjia and Wu, Hongzhi and Zhou, Kun and Jiang, Chenfanfu and others},
  booktitle={Proceedings of the Computer Vision and Pattern Recognition Conference},
  pages={518--529},
  year={2025}
}

@article{ye2024superanimal,
  title={SuperAnimal pretrained pose estimation models for behavioral analysis},
  author={Ye, Shaokai and Filippova, Anastasiia and Lauer, Jessy and Schneider, Steffen and Vidal, Maxime and Qiu, Tian and Mathis, Alexander and Mathis, Mackenzie Weygandt},
  journal={Nature communications},
  volume={15},
  number={1},
  pages={5165},
  year={2024},
  publisher={Nature Publishing Group UK London}
}

@article{liao2022kitti,
  title={Kitti-360: A novel dataset and benchmarks for urban scene understanding in 2d and 3d},
  author={Liao, Yiyi and Xie, Jun and Geiger, Andreas},
  journal={IEEE Transactions on Pattern Analysis and Machine Intelligence},
  volume={45},
  number={3},
  pages={3292--3310},
  year={2022},
  publisher={IEEE}
}

@inproceedings{caesar2020nuscenes,
  title={nuscenes: A multimodal dataset for autonomous driving},
  author={Caesar, Holger and Bankiti, Varun and Lang, Alex H and Vora, Sourabh and Liong, Venice Erin and Xu, Qiang and Krishnan, Anush and Pan, Yu and Baldan, Giancarlo and Beijbom, Oscar},
  booktitle={Proceedings of the IEEE/CVF conference on computer vision and pattern recognition},
  pages={11621--11631},
  year={2020}
}

@inproceedings{silberman2012indoor,
  title={Indoor segmentation and support inference from rgbd images},
  author={Silberman, Nathan and Hoiem, Derek and Kohli, Pushmeet and Fergus, Rob},
  booktitle={European conference on computer vision},
  pages={746--760},
  year={2012},
  organization={Springer}
}

@article{ravoor2020deep,
  title={Deep learning methods for multi-species animal re-identification and tracking--a survey},
  author={Ravoor, Prashanth C and Sudarshan, TSB},
  journal={Computer Science Review},
  volume={38},
  pages={100289},
  year={2020},
  publisher={Elsevier}
}

@article{beeryMegadetector2019,
  author       = {Sara Beery and
                  Dan Morris and
                  Siyu Yang},
  title        = {Efficient Pipeline for Camera Trap Image Review},
  journal      = {CoRR},
  volume       = {abs/1907.06772},
  year         = {2019},
  url          = {http://arxiv.org/abs/1907.06772},
  eprinttype    = {arXiv},
  eprint       = {1907.06772},
  bibsource    = {dblp computer science bibliography, https://dblp.org}
}

@inproceedings{zhang2022bytetrack,
  title={Bytetrack: Multi-object tracking by associating every detection box},
  author={Zhang, Yifu and Sun, Peize and Jiang, Yi and Yu, Dongdong and Weng, Fucheng and Yuan, Zehuan and Luo, Ping and Liu, Wenyu and Wang, Xinggang},
  booktitle={European conference on computer vision},
  pages={1--21},
  year={2022},
  organization={Springer}
}

@inproceedings{wang2023videomae,
  title={Videomae v2: Scaling video masked autoencoders with dual masking},
  author={Wang, Limin and Huang, Bingkun and Zhao, Zhiyu and Tong, Zhan and He, Yinan and Wang, Yi and Wang, Yali and Qiao, Yu},
  booktitle={Proceedings of the IEEE/CVF conference on computer vision and pattern recognition},
  pages={14549--14560},
  year={2023}
}

@misc{LivoxAviaUser,
  title = {Livox {{Avia User Manual}}}
}

@misc{LivoxMid70User2021,
  title = {Livox {{Mid-70 User Manual}}},
  year = 2021,
  month = feb
}

@misc{LivoxTELE15User,
  title = {Livox {{TELE-15 User Manual}}}
}

@misc{lila_science_2026,
  author       = {{LILA Sciences}},
  title        = {{LILA Sciences: Labeled Information Library of Alexandria}},
  year         = {2026},
  url          = {https://lila.science/},
  note         = {Accessed: 2026-03-03}
}

@article{gadotCropNotCrop2024,
  title = {To Crop or Not to Crop: {{Comparing}} Whole-image and Cropped Classification on a Large Dataset of Camera Trap Images},
  shorttitle = {{{SpeciesNet}}},
  author = {Gadot, Tomer and Istrate, {\textcommabelow S}tefan and Kim, Hyungwon and Morris, Dan and Beery, Sara and Birch, Tanya and Ahumada, Jorge},
  year = 2024,
  month = oct,
  doi = {10.1049/cvi2.12318},
  urldate = {2025-09-10},
  langid = {english}
}

@article{valIdentifyingOcelotsIndividual2024,
  title = {Identifying Ocelots by Individual Coat Pattern Marks Using the {{Whiskerbook}}},
  author = {do Val, Helena Gurj{\~a}o Pinheiro and Pezzuti, Juarez Carlos Brito and {Carvalho-Jr}, Elildo Alves Ribeiro},
  year = 2024,
  month = sep,
  journal = {Mammalogy Notes},
  volume = {10},
  number = {2},
  pages = {413--413},
  issn = {2382-3704},
  doi = {10.47603/mano.v10n2.413},
  urldate = {2026-03-04}
}

@InProceedings{Zuffi_2024_CVPR,
    author    = {Zuffi, Silvia and Mellbin, Ylva and Li, Ci and Hoeschle, Markus and Kjellstr\"om, Hedvig and Polikovsky, Senya and Hernlund, Elin and Black, Michael J.},
    title     = {VAREN: Very Accurate and Realistic Equine Network},
    booktitle = {Proceedings of the IEEE/CVF Conference on Computer Vision and Pattern Recognition (CVPR)},
    month     = {June},
    year      = {2024},
    pages     = {5374-5383}
}

@article{ceballosBiologicalAnnihilationOngoing2017,
  title = {Biological Annihilation via the Ongoing Sixth Mass Extinction Signaled by Vertebrate Population Losses and Declines},
  author = {Ceballos, Gerardo and Ehrlich, Paul R. and Dirzo, Rodolfo},
  year = 2017,
  month = jul,
  journal = {Proceedings of the National Academy of Sciences},
  volume = {114},
  number = {30},
  pages = {E6089-E6096},
  publisher = {Proceedings of the National Academy of Sciences},
  doi = {10.1073/pnas.1704949114},
  urldate = {2026-05-10}
}

@article{tuiaPerspectivesMachineLearning2022,
  title = {Perspectives in Machine Learning for Wildlife Conservation},
  author = {Tuia, Devis and Kellenberger, Benjamin and Beery, Sara and Costelloe, Blair R. and Zuffi, Silvia and Risse, Benjamin and Mathis, Alexander and Mathis, Mackenzie W. and {van Langevelde}, Frank and Burghardt, Tilo and Kays, Roland and Klinck, Holger and Wikelski, Martin and Couzin, Iain D. and {van Horn}, Grant and Crofoot, Margaret C. and Stewart, Charles V. and {Berger-Wolf}, Tanya},
  year = 2022,
  month = feb,
  journal = {Nature Communications},
  volume = {13},
  number = {1},
  pages = {792},
  publisher = {Nature Publishing Group},
  issn = {2041-1723},
  doi = {10.1038/s41467-022-27980-y},
  urldate = {2023-06-15},
  copyright = {2022 The Author(s)},
  langid = {english}
}

@article{daviesAdvancesAnimalEcology2014,
  title = {Advances in Animal Ecology from {{3D-LiDAR}} Ecosystem Mapping},
  author = {Davies, Andrew B. and Asner, Gregory P.},
  year = 2014,
  month = dec,
  journal = {Trends in Ecology \& Evolution},
  volume = {29},
  number = {12},
  pages = {681--691},
  issn = {0169-5347},
  doi = {10.1016/j.tree.2014.10.005},
  urldate = {2022-09-13},
  langid = {english}
}

@misc{shuklaWildLIFTLiftingMonocular2026,
  title = {{{WildLIFT}}: {{Lifting}} Monocular Drone Video to {{3D}} for Species-Agnostic Wildlife Monitoring},
  shorttitle = {{{WildLIFT}}},
  author = {Shukla, Vandita and Remondino, Fabio and Costelloe, Blair and Risse, Benjamin},
  year = 2026,
  month = apr,
  number = {arXiv:2604.24718},
  eprint = {2604.24718},
  primaryclass = {cs},
  publisher = {arXiv},
  doi = {10.48550/arXiv.2604.24718},
  urldate = {2026-05-01},
  archiveprefix = {arXiv}
}

@article{kanopoulosDesignImageEdge1988,
  title = {Design of an Image Edge Detection Filter Using the {{Sobel}} Operator},
  author = {Kanopoulos, N. and Vasanthavada, N. and Baker, R.L.},
  year = 1988,
  month = apr,
  journal = {IEEE Journal of Solid-State Circuits},
  volume = {23},
  number = {2},
  pages = {358--367},
  issn = {1558-173X},
  doi = {10.1109/4.996},
  urldate = {2026-05-10}
}

@article{fischlerRandomSampleConsensus1981,
  title = {Random Sample Consensus: A Paradigm for Model Fitting with Applications to Image Analysis and Automated Cartography},
  shorttitle = {Random Sample Consensus},
  author = {Fischler, Martin A. and Bolles, Robert C.},
  year = 1981,
  month = jun,
  journal = {Commun. ACM},
  volume = {24},
  number = {6},
  pages = {381--395},
  issn = {0001-0782},
  doi = {10.1145/358669.358692},
  urldate = {2026-05-10}
}

@article{mathisDeepLearningTools2020,
  title = {Deep Learning Tools for the Measurement of Animal Behavior in Neuroscience},
  author = {Mathis, Mackenzie Weygandt and Mathis, Alexander},
  year = 2020,
  month = feb,
  journal = {Current Opinion in Neurobiology},
  series = {Neurobiology of {{Behavior}}},
  volume = {60},
  pages = {1--11},
  issn = {0959-4388},
  doi = {10.1016/j.conb.2019.10.008},
  urldate = {2022-02-26},
  langid = {english}
}

@article{aamirRobustMetricDistance2025,
  title = {A Robust Metric Distance and Height Estimation Pipeline for Wildlife Camera Trap Imagery},
  author = {Aamir, Muhammad and Wijers, Matthew and Loveridge, Andrew and Markham, Andrew},
  year = 2025,
  month = dec,
  journal = {Ecological Informatics},
  volume = {92},
  pages = {103520},
  issn = {1574-9541},
  doi = {10.1016/j.ecoinf.2025.103520},
  urldate = {2026-05-11}
}

@inproceedings{long2021radar,
  title={Radar-camera pixel depth association for depth completion},
  author={Long, Yunfei and Morris, Daniel and Liu, Xiaoming and Castro, Marcos and Chakravarty, Punarjay and Narayanan, Praveen},
  booktitle={Proceedings of the IEEE/CVF Conference on Computer Vision and Pattern Recognition (CVPR)},
  year={2021}
}

@inproceedings{lin2020depth,
  title={Depth estimation from monocular images and sparse radar data},
  author={Lin, Juan-Ting and Dai, Dengxin and Van Gool, Luc},
  booktitle={IEEE/RSJ International Conference on Intelligent Robots and Systems (IROS)},
  year={2020}
}

@inproceedings{nabati2021centerfusion,
  title={CenterFusion: Center-based radar and camera fusion for 3D object detection},
  author={Nabati, Ramin and Qi, Hairong},
  booktitle={Proceedings of the IEEE/CVF Winter Conference on Applications of Computer Vision (WACV)},
  year={2021}
}

@article{palffy2022multi,
  title={Multi-class road user detection with 3+1D radar in the View-of-Delft dataset},
  author={Palffy, Andras and Pool, Ewoud and Baratam, Srimannarayana and Kooij, Julian F P and Gavrila, Dariu M},
  journal={IEEE Robotics and Automation Letters},
  volume={7},
  number={2},
  pages={4961--4968},
  year={2022}
}

@inproceedings{paek2022kradar,
  title={K-Radar: 4D radar object detection for autonomous driving in various weather conditions},
  author={Paek, Dong-Hee and Kong, Seung-Hyun and Wijaya, Kevin Tirta},
  booktitle={Advances in Neural Information Processing Systems (NeurIPS)},
  year={2022}
}


%% file: v2_additions.bib
@inproceedings{bewley2016sort,
  title     = {Simple Online and Realtime Tracking},
  author    = {Bewley, Alex and Ge, Zongyuan and Ott, Lionel and Ramos, Fabio and Upcroft, Ben},
  booktitle = {IEEE International Conference on Image Processing (ICIP)},
  pages     = {3464--3468},
  year      = {2016}
}

@inproceedings{wojke2017deepsort,
  title     = {Simple Online and Realtime Tracking with a Deep Association Metric},
  author    = {Wojke, Nicolai and Bewley, Alex and Paulus, Dietrich},
  booktitle = {IEEE International Conference on Image Processing (ICIP)},
  pages     = {3645--3649},
  year      = {2017}
}

@article{luiten2021hota,
  title   = {{HOTA}: A Higher Order Metric for Evaluating Multi-Object Tracking},
  author  = {Luiten, Jonathon and O{\v{s}}ep, Aljo{\v{s}}a and Dendorfer, Patrick and Torr, Philip and Geiger, Andreas and Leal-Taix{\'e}, Laura and Leibe, Bastian},
  journal = {International Journal of Computer Vision},
  volume  = {129},
  pages   = {548--578},
  year    = {2021}
}

@article{bernardin2008clearmot,
  title   = {Evaluating Multiple Object Tracking Performance: The {CLEAR MOT} Metrics},
  author  = {Bernardin, Keni and Stiefelhagen, Rainer},
  journal = {EURASIP Journal on Image and Video Processing},
  volume  = {2008},
  pages   = {246309},
  year    = {2008}
}

@inproceedings{weng2020ab3dmot,
  title     = {{3D} Multi-Object Tracking: A Baseline and New Evaluation Metrics},
  author    = {Weng, Xinshuo and Wang, Jianren and Held, David and Kitani, Kris},
  booktitle = {IEEE/RSJ International Conference on Intelligent Robots and Systems (IROS)},
  year      = {2020}
}

@inproceedings{yin2021centerpoint,
  title     = {Center-based {3D} Object Detection and Tracking},
  author    = {Yin, Tianwei and Zhou, Xingyi and Kr{\"a}henb{\"u}hl, Philipp},
  booktitle = {IEEE/CVF Conference on Computer Vision and Pattern Recognition (CVPR)},
  year      = {2021}
}

@inproceedings{yang2020radarnet,
  title     = {{RadarNet}: Exploiting Radar for Robust Perception of Dynamic Objects},
  author    = {Yang, Bin and Guo, Runsheng and Liang, Ming and Casas, Sergio and Urtasun, Raquel},
  booktitle = {European Conference on Computer Vision (ECCV)},
  year      = {2020}
}

@inproceedings{kellner2013instantaneous,
  title     = {Instantaneous Ego-Motion Estimation Using {Doppler} Radar},
  author    = {Kellner, Dominik and Barjenbruch, Michael and Klappstein, Jens and Dickmann, J{\"u}rgen and Dietmayer, Klaus},
  booktitle = {IEEE International Conference on Intelligent Transportation Systems (ITSC)},
  year      = {2013}
}

@article{chen2006microdoppler,
  title   = {Micro-{Doppler} Effect in Radar: Phenomenon, Model, and Simulation Study},
  author  = {Chen, Victor C. and Li, Fayin and Ho, Shen-Shyang and Wechsler, Harry},
  journal = {IEEE Transactions on Aerospace and Electronic Systems},
  volume  = {42},
  number  = {1},
  pages   = {2--21},
  year    = {2006}
}

@inproceedings{farneback2003two,
  title     = {Two-Frame Motion Estimation Based on Polynomial Expansion},
  author    = {Farneb{\"a}ck, Gunnar},
  booktitle = {Scandinavian Conference on Image Analysis (SCIA)},
  series    = {LNCS},
  volume    = {2749},
  pages     = {363--370},
  year      = {2003}
}

@article{kalman1960new,
  title   = {A New Approach to Linear Filtering and Prediction Problems},
  author  = {Kalman, Rudolph E.},
  journal = {Journal of Basic Engineering},
  volume  = {82},
  number  = {1},
  pages   = {35--45},
  year    = {1960}
}

@book{buckland2001distance,
  title     = {Introduction to Distance Sampling: Estimating Abundance of Biological Populations},
  author    = {Buckland, Stephen T. and Anderson, David R. and Burnham, Kenneth P. and Laake, Jeffrey L. and Borchers, David L. and Thomas, Len},
  publisher = {Oxford University Press},
  year      = {2001}
}

@article{howe2017distance,
  title   = {Distance Sampling with Camera Traps},
  author  = {Howe, Eric J. and Buckland, Stephen T. and Despr{\'e}s-Einspenner, Marie-Lyne and K{\"u}hl, Hjalmar S.},
  journal = {Methods in Ecology and Evolution},
  volume  = {8},
  number  = {11},
  pages   = {1558--1565},
  year    = {2017}
}

@article{mathis2018deeplabcut,
  title   = {{DeepLabCut}: Markerless Pose Estimation of User-Defined Body Parts with Deep Learning},
  author  = {Mathis, Alexander and Mamidanna, Pranav and Cury, Kevin M. and Abe, Taiga and Murthy, Venkatesh N. and Mathis, Mackenzie Weygandt and Bethge, Matthias},
  journal = {Nature Neuroscience},
  volume  = {21},
  pages   = {1281--1289},
  year    = {2018}
}

@article{romero2019idtracker,
  title   = {idtracker.ai: Tracking All Individuals in Small or Large Collectives of Unmarked Animals},
  author  = {Romero-Ferrero, Francisco and Bergomi, Mattia G. and Hinz, Robert C. and Heras, Francisco J. H. and de Polavieja, Gonzalo G.},
  journal = {Nature Methods},
  volume  = {16},
  pages   = {179--182},
  year    = {2019}
}

@inproceedings{aamirWildDepth2026,
  title     = {{WildDepth}: A Multimodal Dataset for {3D} Wildlife Perception and Depth Estimation},
  author    = {Aamir, Muhammad and Muramatsu, Naoya and Shin, Sangyun and Wijers, Matthew and Zhong, Jia-Xing and Hou, Xinyu and Patel, Amir and Loveridge, Andrew and Markham, Andrew},
  booktitle = {European Conference on Computer Vision (ECCV)},
  year      = {2026}
}

@inproceedings{ke2024marigold,
  title     = {Repurposing Diffusion-Based Image Generators for Monocular Depth Estimation},
  author    = {Ke, Bingxin and Obukhov, Anton and Huang, Shengyu and Metzger, Nando and Daudt, Rodrigo Caye and Schindler, Konrad},
  booktitle = {IEEE/CVF Conference on Computer Vision and Pattern Recognition (CVPR)},
  year      = {2024}
}
